\documentclass[11pt,a4paper]{article}

\usepackage{fontspec}
\usepackage{unicode-math}
\usepackage{amsmath}
\usepackage{amsthm}
\usepackage{mathtools}

\usepackage[margin=1in,top=1in,bottom=1in]{geometry}
\usepackage{setspace}
\usepackage{enumitem}
\setlist[itemize]{leftmargin=*,topsep=4pt,itemsep=2pt}
\setlist[enumerate]{leftmargin=*,topsep=4pt,itemsep=2pt}

\usepackage{booktabs}
\usepackage{array}
\usepackage{longtable}
\usepackage{tabularx}
\usepackage{graphicx}
\usepackage{float}
\usepackage[skip=4pt,font=small]{caption}

\usepackage{titlesec}
\titleformat{\section}{\normalfont\Large\bfseries}{\thesection}{0.6em}{}
\titleformat{\subsection}{\normalfont\large\bfseries}{\thesubsection}{0.5em}{}
\titlespacing*{\section}{0pt}{1.5ex plus 0.5ex minus 0.2ex}{1ex plus 0.2ex}
\titlespacing*{\subsection}{0pt}{1.2ex plus 0.3ex minus 0.2ex}{0.6ex plus 0.2ex}

\usepackage[hidelinks,colorlinks=false]{hyperref}

\theoremstyle{plain}
\newtheorem{theorem}{Theorem}
\newtheorem{proposition}{Proposition}
\newtheorem{lemma}{Lemma}
\newtheorem{corollary}{Corollary}
\theoremstyle{definition}
\newtheorem{definition}{Definition}
\theoremstyle{remark}
\newtheorem{remark}{Remark}
\newtheorem{assumption}{Assumption}

\renewenvironment{abstract}
  {\small\noindent\textbf{Abstract.}\hspace{0.5em}}
  {\par\medskip}

\title{\textbf{BEHAVE: Human Systems as Observable Complex Dynamical Systems for Physical AI}\\[0.3em]
\normalsize\itshape A Relational, Operational and Dynamical Formalization of Human Systems}

\author{Helene Malyutina\\
\normalsize\itshape BEHAVE Collective Dynamics Lab for Physical AI}

\date{2026 \\ \small\itshape Preprint v2}

\begin{document}
\maketitle

\begin{abstract}
A robot can track every person in a room and still fail to see the system those people form. BEHAVE formalizes that missing level as a \emph{HumanSystem}: an observable, persistent, relational dynamical object whose state is carried partly by interaction structure and is therefore not explicit in independent individual-track representations or reducible to simple aggregates. We call this operational emergence. Public pedestrian data support the premise: on a benchmark with annotated social groups, interaction evidence improves group discrimination beyond proximity (AUC $0.896\to0.933$); on 24 bottleneck runs, future-calm and future-breakdown moments matched on density, mean speed, flow and speed dispersion differ in neighbour-level organization under run-level inference ($p=0.028$).

From observable interaction evidence $K$, BEHAVE constructs conservative routing $P$ and local dynamics $J=-D+GP$, separating routing from gain and relaxation. This makes stability, critical modes and response explicit model quantities, yields exact bounds on what topology can change in collective stability, and gives conditions under which an observable is blind to the mode becoming unstable. In a causal real-data stress test, the coupled operator improves held-out local dynamics over self-only relaxation by $4.5\%$. The fitted stability margin $\mathrm{St}$ is prospectively associated with future throughput loss but overlaps with lag-1 autocorrelation and self-only relaxation, so we interpret it as a structured model-based early-warning quantity rather than a superior scalar alarm.

For Physical AI, the HumanSystem provides the human-side object between perception and action. A robot or scheduler can query group state, structure, critical modes and forced response; action-conditioned stability changes are exposed only when supported by identified changes in human dynamics or signed human--machine coupling. The same formalism extends to mixed human--machine systems through a joint Jacobian.
\end{abstract}


% =====================================================================
\section{Introduction}
\label{sec:intro}
% =====================================================================

A machine can know exactly where every person is and still have no representation of the group as a system.

Perception stacks are built around individuals. They produce detections, poses, tracks and predicted trajectories, and when a collective view is needed they add aggregates over those individuals: occupancy, density, flow, risk or cost maps. Both levels are useful, and neither contains an explicit object for the interacting group itself. Whatever state belongs to the group rather than to its members has nowhere to live in such a representation.

BEHAVE starts from the claim that such state exists and is operationally relevant. When people interact, part of what the group is doing, and of what it will do next, is carried by the relations among them: who is coupled to whom, how strongly, and how that structure is changing. This information is not explicit in any independent individual track and is not recovered by simple sums or averages over tracks. Similar spatial configurations can correspond to a queue, a conversation, a coordinated work cell or a fragile bottleneck because the system-level difference can lie in the interaction pattern.

We call this operational emergence and use the term in a testable sense: two scenes can be equivalent under a declared individual or aggregate representation and still be different system states because their relational organization differs and their subsequent evolution differs. If this holds, an interacting human collective should be treated as what it operationally is: a coupled dynamical system whose state is expressed partly in its interaction structure. A machine that can observe that structure can extract properties of the group as properties of the system, represent them in an explicit operational object, and use that object when reasoning about action, response and transition. Complexity here does not mean chaos or strong nonlinearity; it means that the useful state and response of the whole depend on how its parts are coupled.

The evidence comes from several independent tests rather than a single statistic. DIAMOR tests whether an observable interaction construction recovers real social grouping beyond proximity. J\"ulich asks whether moments matched on density, mean speed, flow and speed dispersion can already differ in neighbour-level organization before different futures. Perturbation datasets test whether disturbances propagate through the collective. A separate J\"ulich operator stress test asks whether an identified structured local operator carries prospective information about downstream throughput once current speed, speed dispersion and throughput are controlled. Restricted BEHAVE recordings test the same object on real internally collected social scenes. Each experiment validates or stress-tests a different layer of the construction.

Adding more per-person features alone does not address this. If the missing information has a persistent owner, changes over time, responds to perturbation and constrains future evolution, the right abstraction is a stateful system object.

BEHAVE introduces that object as a HumanSystem. A HumanSystem is neither the list of people in view nor a label attached to a crowd; it is a persistent, observable, relational operational object reconstructed from measured interaction. Its HumanSystemState records what that system is like now: organization, coordination, momentum, spatial and relational structure, observability and confidence, and, where local dynamics are identified, stability and critical geometry.

The computational move is from
\begin{center}
people $\to$ tracks $\to$ aggregate statistics
\end{center}
to
\begin{center}
people $+$ interaction $\to$ HumanSystem $\to$ HumanSystemState $\to$ dynamics $\to$ response / transition / outcome.
\end{center}

The next move is dynamical. BEHAVE does not stop at an interaction graph. From observable evidence $K$ it constructs a relative topology $P$ and writes local dynamics $J=-D+GP$. $P$ answers where a receiver's coupling is routed; $G$ answers how much dynamical gain the selected state has; $D$ answers how quickly that state relaxes. Keeping them separate prevents sensing confidence from being mistaken for physical gain.

The row-normalization of active rows of $P$ has an important consequence. $P$ is conservative routing: it can redirect or concentrate existing susceptibility, but it cannot manufacture gain that is absent from every actor. Isolated actors have zero topology rows and therefore relax at their own $d_i$. The model predicts when topology can change stability and when it cannot. If every interacting actor is locally subcritical, routing alone cannot make that active subsystem unstable; the relational stability regime is the heterogeneous one, where topology determines whether locally susceptible actors couple into a system-level mode. Stability is then a joint property of topology and actor dynamics.

That distinction sharpens the scientific question. `Relations carry state' does not mean that topology alone determines every system property. It means that relational organization is part of the state-bearing object, and that some quantities (propagation paths, group structure, non-leading modes, transient amplification, response routes and, in heterogeneous regimes, the leading stability mode itself) depend on how the system is connected.

Once the HumanSystem has a Jacobian, new questions become well posed: Which collective mode is slowest? Which actors carry that mode? Which edge or candidate intervention moves it? Will a disturbance decay, spread or transiently amplify? Which observable can see the fragile direction, and which is blind to it? These are system-level questions that are not explicit in independent per-person track representations.

Section~\ref{sec:theorem} connects that dynamical object to measurement. Critical slowing itself is classical. Here, an explicitly constructed HumanSystem Jacobian is mapped to a critical collective mode and then to measurement visibility. The coefficient $C$ in the leading $C/\mathrm{St}$ law depends on the critical left/right modes, the forcing and the observable. A measurement can therefore respond strongly to one kind of change and still be blind to the mode that is becoming slow.

The empirical program has two complementary data layers. Public third-party corpora provide reproducible tests of specific claims and baselines. In parallel, BEHAVE is building a restricted deployment corpus from consented real-world events, used for field auditing, normalization/calibration work and deployment development. The raw footage is not publicly redistributed because consent, privacy/re-identification risk and proprietary calibration constraints differ from those of benchmark trajectory data; the scientific status of restricted results is therefore stated separately and not used as a substitute for public evidence.

For Physical AI, the consequence is concrete. A robot path, AMR motion, scheduler decision or machine cycle is an intervention into a human system. Collision-free trajectory prediction answers whether the machine can move through people; a HumanSystem representation addresses a different question: what system state will that action perturb, which groups or modes will it expose, and whether a candidate action changes the system's response or stability.

Section~\ref{sec:operational-object} makes the action interface executable while keeping human routing normalized independently of robot input: uncalibrated actions enter as external forcing and critical-mode loading, and action-conditioned $\Delta\mathrm{St}(a)$ is exposed only when changes in the human operator are separately identified.

\begin{quote}
\small\itshape
\textbf{Central result.} An interacting human collective carries operational state in its interaction structure, and this state is not explicit in independent individual-track representations and is not reducible to simple aggregates over them. BEHAVE makes that state explicit and computable: it defines the HumanSystem as a persistent, observable operational object; gives it local dynamics that separate interaction evidence, routing, gain and relaxation; states exactly when topology can and cannot change collective stability; and connects the system's critical modes to what a sensor can and cannot see. Public data support the premise that relational organization carries state beyond the tested individual and flow-level descriptions, and a causal real-data stress test shows that the resulting structured operator carries prospective information on real crowds while overlapping with classical critical-slowing indicators. For Physical AI, this turns the human side of a shared workspace from a collection of trajectories into a system a machine can query, compare actions against and, where dynamics are identifiable, compose with its own dynamics. Graphs, linear systems and critical slowing are established tools. BEHAVE combines them in an architecture that represents interacting humans as a measurable, stateful and actionable system object.
\end{quote}

\subsection{Contributions}
\label{sec:contributions}

\begin{enumerate}
\item[\textbf{C1}] \textbf{Relational organization is empirically state-bearing.} Multiple public datasets support distinct parts of the same representational claim: observable interaction beyond proximity carries group information; J\"ulich states matched on density, mean speed, flow and speed dispersion retain a run-level difference in neighbour-speed organization before different futures; perturbations propagate through human collectives; and causal collective-state signals can move before downstream consequences. These results establish information beyond the tested individual/flow descriptions without claiming irreducibility relative to every conceivable statistic.
\item[\textbf{C2}] \textbf{HumanSystem as an operational ontology.} BEHAVE defines a persistent HumanSystem distinct from its actors and from its current HumanSystemState, with explicit objects for interaction evidence, topology, groups, boundaries, transitions, perturbations, responses, outcomes and observability. The ontology is machine-readable and implemented in the engine.
\item[\textbf{C3}] \textbf{Interaction becomes dynamics through a conservative-routing decomposition.} BEHAVE separates observable coupling evidence $K$, conservative routing $P$, dynamical gain $G$ and intrinsic relaxation $D$, yielding $J=-D+GP$. Active topology rows are stochastic; isolated actors have zero rows. This produces row- and column-sum bounds on the spectral margin: routing cannot manufacture gain, while gain-weighted incoming topology supplies a topology-dependent envelope and edge sensitivity resolves which feasible reroutings actually move the critical mode. In real J\"ulich windows, sensitivity of $\mathrm{St}$ to replacing $P$ by uniform routing increases with fitted $D/G$ heterogeneity across all tested interaction bandwidths.
\item[\textbf{C4}] \textbf{Critical geometry becomes observable.} For the local stochastic model, the leading $C/\mathrm{St}$ behavior of a quadratic observable is characterized with $C$ in closed form from the critical mode, diffusion and measurement geometry. This provides a HumanSystem-specific bridge from measured interaction to critical collective mode to sensor visibility, including exact blindness conditions; the homogeneous consensus limit gives a concrete example in which graph discord is blind to the critical mode.
\item[\textbf{C5}] \textbf{The operator layer has a causal real-data stress test against classical early-warning baselines.} On the same 24 J\"ulich runs, a trailing-window $D/G$ fit with observed routing improves held-out local dynamics over self-only relaxation. The fitted spectral margin tracks future throughput, while self-only relaxation and lag-1 autocorrelation are also predictive; the coupled margin contributes a modest residual under several controls but is not claimed to dominate classical early-warning statistics. Direct unconstrained LIM fails the longer-lag $\tau$-consistency test, bounding the local linear-Markov regime.
\item[\textbf{C6}] \textbf{Executable human-side systems layer for Physical AI.} Candidate machine actions map first to explicit external forcing and critical-mode loading. An action-conditioned $\Delta\mathrm{St}$ is exposed only when data support a corresponding change in $D$, $G$, human routing $P_{HH}$, or signed human--machine coupling; uncalibrated robot proximity is not allowed to alter the internal margin by normalization alone.
\end{enumerate}

% =====================================================================
\section{From Individuals to a Human System}
\label{sec:why-humansystem}
% =====================================================================

The central question is whether a machine loses operationally relevant state when it represents a scene only as individuals plus aggregate statistics.

If the answer were no, there would be no need for BEHAVE. Better tracking, better trajectory prediction and better aggregation would be sufficient. The evidence says the answer is yes in the settings studied here.

\subsection{Relations carry information that the conventional view can miss}

DIAMOR provides the first piece. Two people being close is not the same as two people forming an interacting unit. Adding observable directional interaction to a proximity baseline improves group-membership discrimination on real trajectories (DIAMOR-1 AUC $0.8955\to0.9331$, replicated on a second recording day). What matters here is less the absolute AUC than the finding that an interaction-aware representation contains information that a geometry-only representation discards.

J\"ulich provides the second and conceptually stronger piece. The original public matched-state test compares local bottleneck moments matched on density, mean speed and flow. We reran the analysis on all 24 uploaded 2021 bottleneck trajectories with speed dispersion added as a fourth matching variable and with inference reduced to one effect per experimental run. Nineteen runs yielded informative matched pairs (3639 pairs total). Future-breakdown states retained higher nearest-neighbour speed disagreement: mean run-level delta $=+0.00449\,$m/s, 12/19 runs positive, exact run-level sign-flip $p=.0277$, run-bootstrap $95\%$ CI $[+0.00033,+0.00872]$. Thus the effect is not explained simply by local speed dispersion. A stricter sensitivity control that subtracts the expected neighbour disagreement under random reassignment of the exact same speeds to the exact same positions remains suggestive rather than decisive (sign-flip $p=.0459$; bootstrap CI crosses zero; Wilcoxon $p=.072$).

This is the operational meaning of emergent state used throughout the paper: the relational organization can distinguish system states that disappear when the representation collapses the collective into independent tracks or aggregate summaries.

\subsection{A collective is not only relational; it is dynamical}

A state is useful only if it participates in evolution. The datasets therefore probe more than static structure. In DIAMOR, an explicit passing perturbation changes group motion. In J\"ulich impulse experiments, disturbances propagate through the human row with measurable delay structure. In the bottleneck runs, a collective state built causally from trailing observables changes before a separately defined throughput degradation in 12 of 13 runs. These are not yet measurements of $\mathrm{St}$. They establish the premise required before spectral modeling makes sense: the collective has temporally persistent, perturbable, state-dependent organization.

The distinction matters. Event detection asks whether something happened. A dynamical representation asks what state the system is in, how that state persists, how it changes under perturbation, and what modes govern its response.

\subsection{The missing object}

Once relations carry state and that state has temporal response, there must be an object to which the state belongs. BEHAVE calls it the HumanSystem.

The HumanSystem is not a psychological construct. It is reconstructed from observables and explicitly derived quantities. It is also not synonymous with ``crowd'': the same object model applies to a small interacting group, a production cell, a queue, a team around a robot or another bounded human system, provided the observability and evaluation conditions of Section~\ref{sec:scope} hold.

The representational progression is therefore:
\begin{center}
individual observables $\to$ interaction evidence $\to$ relational topology and group structure $\to$ HumanSystemState $\to$ explicit local dynamics $\to$ transition / response / operational outcome.
\end{center}
Sections~\ref{sec:ontology}--\ref{sec:fields} formalize each arrow. The empirical sections do not claim that every layer is equally validated today. Instead, they establish the lower layers strongly and give the spectral layer a first causal real-data stress test while keeping full identification and uncertainty quantification explicitly open. The contribution is the systems architecture itself; no single alarm score is proposed.

% =====================================================================
\section{Ontology and Observability}
\label{sec:ontology}
% =====================================================================

\subsection{Core claim}
\label{sec:core-claim}

The core representational claim is this: individuals are observed, relations are reconstructed, and the collective is modeled as a persistent HumanSystem whose state belongs to the system rather than to any one actor. This lets BEHAVE talk about emergence without leaving computation: emergence is the appearance of operationally relevant state at the relational level of description.

The HumanSystem is then given local dynamics, making stability, modes, response and intervention sensitivity questions about a defined model rather than metaphors borrowed from dynamical-systems language. Table~\ref{tab:ontology} fixes the object model used throughout the paper.

\subsection{HumanSystem Operational Ontology}
\label{sec:entities}

The ontology below is an operational object model: the minimal set of computational entities and distinctions that the rest of BEHAVE constructs, reasons about and acts on. It is not a taxonomy of social life, and it is implemented as a machine-readable state object (Section~\ref{sec:implementation}).

\begin{table}[H]
\centering\footnotesize
\begin{tabularx}{\textwidth}{@{}>{\raggedright\arraybackslash}p{3.1cm}>{\raggedright\arraybackslash}X>{\raggedright\arraybackslash}p{1.5cm}@{}}
\toprule
\textbf{Object} & \textbf{Meaning} & \textbf{Built in} \\
\midrule
HumanActor & An observed person participating in the current system. & \S\ref{sec:obs-layer} \\
Observable & A quantity directly supported by measurement: position, velocity, pose or orientation where reliably observable, measurement confidence. & \S\ref{sec:obs-layer} \\
InteractionEvidence $K(t)$ & Absolute evidence that one actor is behaviorally coupled to another now. Not topology, not dynamics. & \S\ref{sec:K-def} \\
InteractionTopology $P(t)$ & The normalized routing of coupling derived from $K$: how coupling is distributed, not how strong it is dynamically. & \S\ref{sec:P-def} \\
InteractionGroup & A structurally supported subgroup of the current system; ungrouped actors are kept explicit. & \S\ref{sec:G-def} \\
Boundary / Bridge & Weak structural separation between groups; connective actors or edges across it. & \S\ref{sec:G-def} \\
SpatialZone & A region of the scene over which spatial aggregates are computed. & \S\ref{sec:fields} \\
HumanSystem $\mathcal H(t)$ & The persistent computational object for the interacting collective, distinct from its actor set, from any single observation, and from its current state. & \S\ref{sec:H-vs-SH} \\
HumanSystemState $S_{\mathcal H}(t)$ & What this HumanSystem is like now: regime, coordination, momentum, relational and spatial structure, group-level state, stability or fragility where identifiable, observability and confidence. Task-dependent; not one universal vector. & \S\ref{sec:SH-def} \\
Dynamics $J_H(t)$ & The local model of evolution around the current operating regime; in the present positive-system specialization $J_H=-D+GP$. & \S\ref{sec:dynamics} \\
StateTransition & A persistent or dynamical change of $S_{\mathcal H}$. A visible event is not a state transition. & \S\ref{sec:state-transitions} \\
Perturbation $u_{ext}$ & An explicitly declared external intervention or disturbance: machine motion, an environmental change, an externally introduced pedestrian, an operational intervention. & \S\ref{sec:state-eq} \\
Response & The state-dependent change of the HumanSystem conditional on a declared perturbation: absorption, amplification, propagation, reorganization, split, merge, recruitment, recovery. & \S\ref{sec:response} \\
OperationalOutcome & A downstream external consequence: throughput degradation, delay, blockage, task failure. A state transition is not an outcome. & \S\ref{sec:empirical} \\
Observability / Confidence & A first-class part of the object: a quantity exists operationally only when its measurement or identification conditions are met. & \S\ref{sec:obs-first} \\
\bottomrule
\end{tabularx}
\caption{HumanSystem Operational Ontology. Fields such as $T$, $S$, $L$, $A$, $N$, $M$ (Section~\ref{sec:fields}) are derived observables that populate $S_{\mathcal H}$; they are not themselves ontology objects.}
\label{tab:ontology}
\end{table}

\textbf{Relations.} \emph{interactsWith}, \emph{memberOf}, \emph{adjacentTo} / \emph{boundedBy}, \emph{occupies}, \emph{perturbedBy}, \emph{respondsTo}, \emph{transitionsTo}, \emph{affects}.

The rest of the paper constructs and uses these objects in order: Section~\ref{sec:interaction} builds $K$, $P$ and group structure; Section~\ref{sec:dynamics} the local dynamics; Section~\ref{sec:theorem} the spectral geometry of those dynamics; Section~\ref{sec:fields} derives components of $S_{\mathcal H}$; Section~\ref{sec:empirical} tests specific premises of the model; and Section~\ref{sec:operational} acts on the HumanSystem through perturbation, response and composition with machine dynamics.

\subsection{HumanSystem versus HumanSystemState}
\label{sec:H-vs-SH}

BEHAVE distinguishes the system as an object from the system's state at a point in time.

\begin{definition}[HumanSystem and HumanSystemState]
\label{def:H-SH}
Let $\mathcal{H}(t)$ denote the \emph{HumanSystem}: the computed relational object consisting of the actor set, the interaction structure among them, and the rules by which further quantities are derived, as constituted at or up to time $t$. Let $S_{\mathcal{H}}(t)$ denote the \emph{HumanSystemState}: the specific value, at time $t$, of the collection of derived quantities (structural, field-level, and dynamical) that $\mathcal{H}(t)$ admits. We maintain $\mathcal{H}(t) \neq S_{\mathcal{H}}(t)$ throughout: $\mathcal{H}(t)$ is the object and its generative structure; $S_{\mathcal{H}}(t)$ is a snapshot of that object's derived properties.
\end{definition}

The separation is used later in two places. Section~\ref{sec:fields} can treat $S_{\mathcal{H}}(t)$ as an open, task-dependent collection of components rather than a fixed-length vector without changing the identity of $\mathcal{H}(t)$. It also makes a \emph{state transition} (Section~\ref{sec:operational}) a well-posed notion distinct from an \emph{event}: a state transition is a change in $S_{\mathcal{H}}(t)$, not merely the occurrence of something visible in the raw signal.

\begin{remark}[Scope of HumanSystem identity]
\label{rem:identity}
Calling $\mathcal{H}(t)$ persistent raises an identity question that this paper answers by scoping rather than by theory: if actors leave and others arrive, or two groups merge and later split, what makes the result the same HumanSystem? We adopt the minimal convention that a HumanSystem's identity is fixed by a \emph{declared scene and task interval} (the sensing region and time window a given deployment names in advance), and that membership, group structure, and boundaries may all vary within that interval without implying that the object has been replaced. Group-level identity within the system (which InteractionGroup is ``the same'' across a merge or split) and cross-interval identity resolution (whether the system observed on Tuesday is the same one observed on Wednesday) are both outside the scope of this paper; where a group's fate matters operationally, it is listed among the open problems of Section~\ref{sec:limitations} rather than resolved by definition here.
\end{remark}

\subsection{Observability-first principle}
\label{sec:obs-first}

\begin{definition}[Observability-first principle]
No property is attributed to an Actor, an InteractionGroup, or the HumanSystem unless it is either (i) a directly measured observable, or (ii) a quantity derived from observables through an explicitly stated, computable rule, with the identifiability of that quantity from the available signals established in advance for the setting in which it is used.
\end{definition}

In particular: motion, proximity, and kinematic micro-signals are not automatically mapped onto emotion, intention, personality, or any other latent psychological label. Derived quantities introduced later in this paper carry names such as ``tension,'' ``attention,'' or ``synchrony''; each such name denotes a specific computable functional of observable kinematics and interaction evidence, defined at the point it is introduced, and carries no further claim about the internal state of any actor unless such a claim is separately argued and tested.

Observability is enforced computationally as well as conceptually. When metric calibration, orientation reliability, dynamical identifiability or deployment calibration is insufficient, the corresponding HumanSystem quantities are withheld rather than replaced by an inferred universal score (Section~\ref{sec:implementation}).

\subsection{Observability layer}
\label{sec:obs-layer}

The observability layer fixes what enters the rest of the pipeline.

\begin{itemize}
\item \textbf{Input.} Video, tracked trajectories, or an equivalent sensing stream over a set of actors in a scene $\Omega \subset \mathbb{R}^p$, $p \in \{2,3\}$, over a time interval.
\item \textbf{Observable outputs.} For each actor $i$ and time $t$: position $x_i(t)$, and, when reliably extractable, velocity $v_i(t)$, pose or orientation, a measurement confidence $c_i(t) \in [0,1]$, and a timestamp. Any quantity not listed here is not assumed available and must be justified before use.
\item \textbf{What is explicitly not assumed observable.} Emotional state, intention, personality, or any other latent psychological attribute of an actor. These are not computed at any stage of the pipeline described in this paper.
\end{itemize}

This layer is the only point at which the model touches raw sensing. Everything from Section~\ref{sec:interaction} onward (interaction evidence, topology, fields, state, dynamics) is defined as a function of this layer's outputs, of quantities derived upstream in the architecture, and of dynamical parameters estimated from them (Section~\ref{sec:identification}), and of nothing added later.

\begin{figure}[H]
\centering
\includegraphics[width=\textwidth]{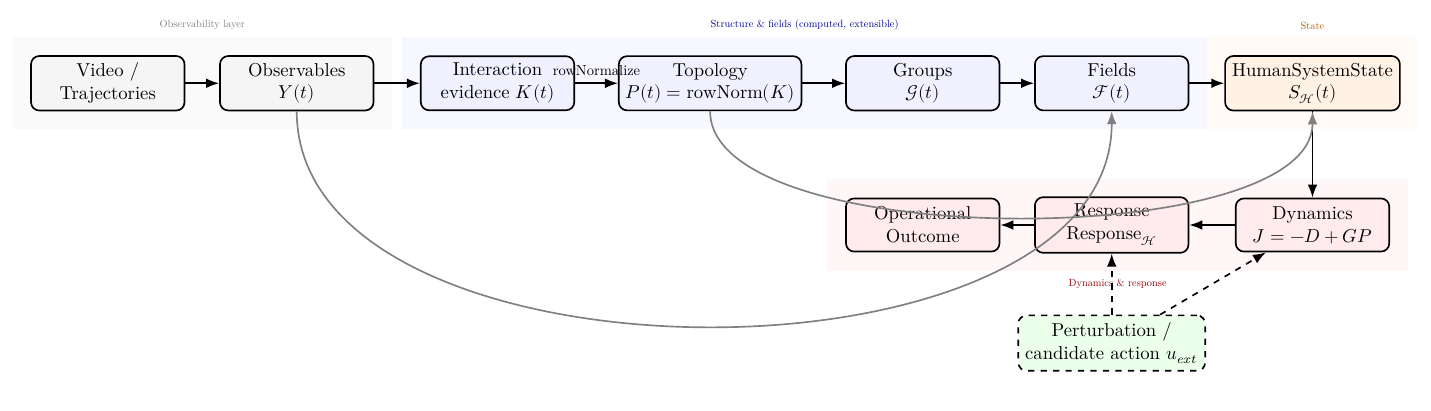}
\caption{The BEHAVE v2 pipeline as a dependency DAG, drawn directly from the architecture invariant above: observability (gray) feeds structure and fields (blue, computed and extensible, not a fixed basis), which feed state (orange), which feeds dynamics, response, and operational outcome (red). The dashed green node is the one explicit external input, $u_{ext}$; it is not derived from the system's own state. Gray arcs mark declared same-invariant shortcuts (e.g. fields may read observables directly); every arrow still points forward in the dependency order.}
\label{fig:architecture}
\end{figure}

\begin{definition}[Architecture invariant]
Every derived quantity in this paper is defined only in terms of the observability layer's outputs together with quantities defined upstream of it in the dependency order $Y \to K \to P \to \mathcal{G} \to \mathcal{F} \to S_{\mathcal H} \to (\text{dynamics}) \to \mathrm{Response} \to \mathrm{OperationalOutcome}$ (Figure~\ref{fig:architecture}). No downstream construct may depend on a hidden label, future information, or an undeclared variable. An explicit external input $u_{ext}$ (a perturbation or a candidate action) is allowed at the dynamics/response stage and must be named where it enters, as in Section~\ref{sec:dynamics}. This is a \emph{dependency} invariant: it fixes what each quantity may be a function of, so that no hidden label or future information enters the architecture. The dynamical parameters $D$ and $G$ sit inside it and are recovered from data through the identification problem of Section~\ref{sec:identification}.
\end{definition}

This is what allows every downstream quantity in this paper to be traced back to a named, observable input, and it is the standard against which Section~\ref{sec:negative}'s negative results are read: a result fails this invariant only if it turns out to depend on something not on this list, which none of the definitions in Sections~\ref{sec:interaction}--\ref{sec:empirical} do.

% =====================================================================
\section{Interaction Structure}
\label{sec:interaction}
% =====================================================================

Interaction is the core information channel of BEHAVE: how the state of one actor is coupled to the states of others, over and above where people are or what each person is doing. This section separates two things that are usually conflated, evidence that a coupling exists and how strong it is ($K$) and how coupling is routed through the system ($P$), and then extracts group structure from the routing.

\subsection{Interaction evidence}
\label{sec:K-def}

\begin{definition}[Interaction evidence]
For actors $i,j$ and time $t$, $K_{ij}(t) \ge 0$ is the observable directed coupling evidence from actor $j$ to actor $i$: a nonnegative functional of declared observables such as spatial proximity, relative motion, persistence of co-presence, orientation (when reliably extractable), and response lag. $K(t) = [K_{ij}(t)]$ is the interaction evidence matrix.
\end{definition}

$K_{ij}(t)$ is an absolute measure of evidence strength: it can be small because two actors are far apart, uncoupled, or simply not both reliably observed at $t$. It does not by itself express how an actor's influence is distributed among its neighbors; that relational question is addressed next.

\subsection{Interaction topology}
\label{sec:P-def}

\begin{definition}[Interaction topology]
\label{def:P}
$$P(t) := \mathrm{rowNormalize}(K(t)), \qquad P_{ij}(t) = \frac{K_{ij}(t)}{\sum_k K_{ik}(t)} \ \ (\text{when } \textstyle\sum_k K_{ik}(t) > 0).$$
$P(t)$ is the relative directed interaction topology: for each $i$ with $S_i(t)=\sum_k K_{ik}(t)>0$, the row $P_{i\cdot}(t)$ distributes actor $i$'s incoming coupling across its sources and sums to one. We use $K_{ii}=P_{ii}=0$. If $S_i(t)=0$, the convention is $P_{i\cdot}(t)=0$: the actor is isolated in the current interaction graph and its topology row contributes no coupling. Thus $P$ is row-substochastic globally and row-stochastic on active interaction rows. Define $\eta_i=\mathbf 1[S_i>0]$ for use in the stability bounds below.
\end{definition}

\begin{remark}[$K \neq P$]
The separation is architectural. $K$ carries absolute evidence strength (it can grow or shrink as a whole, e.g. under a general increase in activity); $P$ carries only relative routing (it is invariant to uniformly rescaling a row of $K$). A model that conflates the two cannot distinguish ``actor $i$ is weakly coupled to everyone'' from ``actor $i$ is strongly coupled, but exclusively to one neighbor.'' Every quantity defined from this point forward that concerns \emph{topology} (connectivity, communities, bridges, structural boundaries, topology change) is a function of $P$; quantities that concern coupling \emph{strength} on top of that topology (Section~\ref{sec:dynamics}) reintroduce magnitude through a separate gain term, not through $K$ directly.
\end{remark}

\subsection{Group structure}
\label{sec:G-def}

Group structure occupies its own position in the architecture invariant, between topology and fields, and is therefore given its own definition rather than left implicit in the diagram.

\begin{definition}[Group structure]
\label{def:group-structure}
$$\mathcal{G}(t) := F_G\big(P(t), Y(t)\big),$$
where $F_G$ is a declared, computable group-extraction rule, fixed for the task before evaluation. $\mathcal{G}(t)$ may contain interaction groups, ungrouped actors, bridges, and structural boundaries.
\end{definition}

BEHAVE does not assume a universal $F_G$, and no claim is made that one exists: different tasks may use different, separately validated group-extraction rules. What the architecture does constrain is the rule's inputs. $F_G$ may depend only on quantities available upstream of it (the observables $Y(t)$ and the topology $P(t)$), and in particular fields are computed \emph{after} $\mathcal{G}(t)$ and may not be used to define the same group structure they are then computed relative to, unless such a dependency is explicitly declared and evaluated as a separate construction. Fields describe the state inside an already-identified structure; using them to define that structure would make the two mutually defining, which is the circularity this ordering exists to prevent.

% =====================================================================
\section{Local Dynamics}
\label{sec:dynamics}
% =====================================================================

The previous sections construct the relational object. This section gives it local dynamics and, just as importantly, states what the chosen normalization allows topology to do.

A graph by itself says who is connected. A dynamical system says what happens when state changes: what relaxes, what propagates, what amplifies, what persists, and which collective mode dominates near a regime boundary. BEHAVE separates observable evidence that coupling exists ($K$), the relative routing of that coupling ($P$), the dynamical gain of the selected state ($G$), and intrinsic relaxation ($D$). That separation is the center of the model.

The interaction Jacobian is $J=-D+GP$. Once a state coordinate and its dynamics are identified, the HumanSystem has a local model whose stability, reactivity, transient gain, critical modes and intervention sensitivity can be computed. The same construction also yields a useful negative statement: row-normalized topology cannot manufacture net gain in a population whose every actor is locally subcritical. Topology matters through routing and through heterogeneity, and the resulting bounds are made explicit below.

This section defines the model, two complementary identification routes, the positive-system diagnostics, and the structural limits that distinguish topology from gain.

\subsection{State equation and the interaction Jacobian}
\label{sec:state-eq}

\begin{definition}[Local dynamics]
\label{def:local-dynamics}
Let $s(t) \in \mathbb{R}^n$ be a local state vector over actors (or, at coarser granularity, over identified groups). The local dynamics are
$$
\dot{s} = -D(t)\,s + G(t)P(t)\,s + B\,u_{ext}(t) + \xi(t),
$$
where $D(t) = \mathrm{diag}(d_1,\dots,d_n)$ collects local relaxation rates, $G(t) = \mathrm{diag}(g_1,\dots,g_n)$ collects local coupling gains, $u_{ext}(t)$ is an explicit external perturbation or candidate action (Section~\ref{sec:ontology}) entering through a loading matrix $B$, and $\xi(t)$ is residual fluctuation. The \emph{interaction Jacobian} is
$$
J(t) := -D(t) + G(t)P(t).
$$
\end{definition}

\begin{remark}[Status of the dynamical model]
\label{rem:by-construction}
BEHAVE constructs an observable relational state $s(t)$, an explicit evolution equation for it, and an interaction Jacobian $J$ whose entries are built from measured interaction evidence. The collective is then a dynamical system in the technical sense: $J$ has a spectrum because it is a matrix in a stated linear system, and stability, reactivity, eigenvalue sensitivity and critical slowing are consequences of the model under their own hypotheses. How well the model, with parameters identified from data, describes a particular collective is then a question answered by measurement.
\end{remark}

$D$, $G$, and $P$ are each independently meaningful: $D$ sets how fast an actor's local state decays absent coupling; $G$ sets how strongly an actor's neighbors' states are transmitted to it; $P$ sets which neighbors and in what relative proportion. Together they determine whether a disturbance entering the collective is absorbed (relaxation dominates), amplified (coupling dominates), or propagated along particular routes of the topology. This moves the representation from a graph of who is near whom toward a dynamical account of how perturbations travel through a human system. $J$ is the object whose spectrum governs local stability and transient behavior; everything in the remainder of this section is a diagnostic computed from $J$.

\begin{remark}[Where $K$'s absolute magnitude goes]
\label{rem:K-magnitude}
Row-normalization discards row-scale magnitude from $P$, and the dynamics above use $P$, not $K$. This is deliberate: the magnitude that matters dynamically is re-entered through $G$, which is estimated from the temporal behavior of $s(t)$ rather than read off the observation kernel. The architectural claim is that \emph{observation evidence is not dynamical gain}: $K_{ij}$ measures how much evidence the sensing layer has that $j$ couples to $i$, which is confounded with visibility, tracking quality, scene geometry, and how long both were observable, whereas $g_i$ is how strongly $i$'s state actually follows its neighbors', which is a property of the dynamics and must be identified from them. Reading $g_i$ off $\sum_j K_{ij}$ would import every sensing artifact in $K$ directly into the system's estimated stability margin. $K$'s absolute scale therefore retains its role in the observability and confidence layer, where it tells us whether a row of $P$ is well estimated at all, and does not enter $J$ except through the separately estimated $G$.
\end{remark}

\begin{remark}[Positive coupling and its signed extension]
\label{rem:signed}
The state coordinate $s_i$ is a declared dynamical state variable, not necessarily a position or velocity, so $GP\ge0$ does not mean that people physically attract or follow one another. It means that an increase in a neighbor's selected state contributes with the same sign to the receiver's state dynamics. Physical avoidance can still propagate positively in coordinates such as deceleration, hesitation, disturbance magnitude, tension-like kinematic intensity or local stopping activity. The Metzler specialization does not represent inhibitory or anti-aligned coupling in the selected coordinate; its natural extension is $J=-D+W$ with signed off-diagonal $W$. In that extension the generic Jacobian quantities ($\mathrm{St}$, reactivity, transient gain, eigenvalue sensitivity, and Theorem~\ref{thm:divergence} under its stated spectral assumptions) remain meaningful. The positive-system results (the $\chi$/M-matrix equivalence of Proposition~\ref{prop:chi-stability} and the Perron positivity of Remark~\ref{rem:perron}) do not carry over automatically, and a leading complex pair (a Hopf-type transition; Kuznetsov, 2004) becomes possible.
\end{remark}

\subsection{Dynamical parameter identification}
\label{sec:identification}

The preceding sections determine the observable and relational part of the architecture: $K$ and $P$ are computed from observables. Passing from that representation to empirical spectral diagnostics requires one additional step: identification of the local dynamical parameters $D$ and $G$ from finite trajectory windows. This section states that step precisely: what must be solved, under what conditions it is solvable, and what makes it hard in practice.

Given a window of trajectories, $P(t)$ is computed directly. What is not directly observed is $s(t)$ together with the pair $(D,G)$. Fixing a declared per-actor state $s_i(t)$ (for the fields of Section~\ref{sec:fields}, a normalized kinematic quantity), the model of Definition~\ref{def:local-dynamics} is linear in the unknown parameters, so with $\dot s$ estimated by finite differences over the window, each actor $i$ contributes rows of a regression
$$
\dot s_i(t) \;=\; -d_i\,s_i(t) \;+\; g_i\,(P(t)s(t))_i \;+\; b_i^\top u_{ext}(t) \;+\; \xi_i(t),
$$
linear in the two unknowns $(d_i,g_i)$ with regressors $s_i(t)$ and $(Ps)_i(t)$.

Two distinct requirements follow. \emph{Structural} identifiability requires the corresponding design matrix to have full rank or, equivalently, sufficient excitation of the self-state and neighbor-state regressors over the window. This fails in recognizable regimes: when an actor's own state and its neighbors' topology-weighted average move together, as in a uniformly synchronized crowd or a near-static configuration where $P$ barely changes, $d_i$ and $g_i$ are not separately identified and only the combination that sets the local decay rate is recoverable. Non-stationarity of $P$ within the window supplies excitation; an explicit perturbation $u_{ext}$ supplies more of it, which is one operational reason to treat perturbations as first-class (Section~\ref{sec:response}).

\emph{Practical} estimation from real trajectories is harder than the regression form suggests, and ordinary least squares is only the idealized noise-free case. $\dot s$ is not observed but differentiated from noisy tracks; $P(t)$ is itself an estimate carrying its own error, so the regressors are noisy, an errors-in-variables problem under which least squares is biased rather than merely imprecise. Residuals are unlikely to be independent across time. A usable estimator therefore needs derivative- and measurement-error-aware system identification with uncertainty quantification (Lehmann \& Casella, 1998), not a least-squares fit.

Two further quantities inherit that conditioning. $\mathrm{St}$ depends on $J$ and therefore on the estimated $(D,G)$; $r_1$ and $l_1$ depend additionally on the separation of $\lambda_1$ from the rest of the spectrum, so a poorly conditioned or nearly degenerate estimated spectrum yields critical modes with wide uncertainty even where $\mathrm{St}$ itself is stable.

Practical use therefore requires a finite-window identification method for $D$ and $G$, with uncertainty propagated to the spectral quantities.

A complementary route is to estimate the local linear operator $J$ directly before decomposing it into $D$, $G$ and $P$. Linear Inverse Modeling (Penland, 1989; Penland \& Sardeshmukh, 1995) uses lagged covariances of a stationary Markov state, $C(\tau)=\mathbb{E}[x(t+\tau)x(t)^\top]$, to obtain an empirical propagator and, when the matrix logarithm is well posed, $J_{LIM}(\tau)=\tau^{-1}\log\!\big(C(\tau)C(0)^{-1}\big)$. The associated tau test asks whether the inferred operator is stable across lag choices and therefore provides a direct diagnostic of the local linear-Markov assumption rather than assuming it silently.

For BEHAVE this gives a strong identification baseline: estimate an unconstrained $J_{LIM}$ from the declared HumanSystem state, compare its held-out dynamics and spectrum with the constrained $J_B=-D+GP$, and only then interpret the constrained factors $D$ and $G$. Agreement would show that the relational factorization captures the empirically inferred local operator; disagreement would localize the failure to state choice, topology construction or model form. Direct-$J$ estimation and constrained $D/G$ estimation are therefore complementary tests, not competing definitions of the HumanSystem.

\subsection{Stability, coupling load, and reactivity}
\label{sec:stability}

\begin{definition}[Stability margin]
\label{def:stability-margin}
$$\mathrm{St}(t) := -\max_i \mathrm{Re}\,\lambda_i(J(t)) \quad [1/\mathrm{s}].$$
$J(t)$ is Hurwitz (locally stable) iff $\mathrm{St}(t) > 0$; $\mathrm{St}(t) \to 0^+$ marks approach to a local stability boundary.
\end{definition}

$\mathrm{St}$ is a distance to a local stability boundary measured in dynamical-system space, not a Euclidean distance between raw measurements: it is the rate at which the slowest collective mode relaxes after a disturbance. As $\mathrm{St}\to0^+$ that mode slows, disturbances along it persist for increasingly long times, fluctuations accumulate preferentially in its direction, and the collective's susceptibility to perturbation grows. This is the critical slowing associated with approaching a bifurcation of the corresponding dynamical regime. ``How close is this group to losing its present regime?'' is, for the model, the question of how small $\mathrm{St}$ is. Section~\ref{sec:theorem} turns this into a quantitative statement about what an observer measures.

\begin{remark}[Perron structure of the HumanSystem Jacobian]
\label{rem:perron}
Because $D$ is diagonal and $G,P\ge0$ entrywise, $J=-D+GP$ is Metzler. Its spectral abscissa is therefore attained by a real eigenvalue (Berman \& Plemmons, 1994): a leading complex-conjugate pair, and with it a Hopf-type first loss of stability, cannot occur in this model class. When $P$ is irreducible and every $g_i>0$, the leading (Perron) eigenvalue is algebraically simple, with entrywise positive right and left critical modes $r_1,l_1$. Reducible coupling (disconnected or one-way-connected subgroups) can instead produce a repeated leading eigenvalue, so the simplicity and gap conditions required by the theorem of Section~\ref{sec:theorem} remain explicit assumptions there.
\end{remark}

\begin{definition}[Dimensionless coupling load]
$$\chi(t) := \rho\!\left(D(t)^{-1}G(t)P(t)\right),$$
the spectral radius of $D^{-1}GP$.
\end{definition}

\begin{proposition}[Coupling load and stability, positive systems]
\label{prop:chi-stability}
If $J(t) = -D(t) + G(t)P(t)$ is Metzler (off-diagonal entries nonnegative, which holds here since $G,P \ge 0$ entrywise and $D$ is diagonal), then $J(t)$ is Hurwitz if and only if $\chi(t) < 1$.
\end{proposition}

This is a standard fact about Metzler/positive linear systems (an $M$-matrix characterization of stability), restated here in BEHAVE's notation because $\chi$ functions as a dimensionless early-warning diagnostic in this architecture, whose empirical status is addressed in Section~\ref{sec:negative}; the stability equivalence itself is not a new result.

\renewcommand{\theproposition}{1A}
\begin{proposition}[Conservative-routing bounds, including isolated actors]
\label{prop:1A}
Let $D=\mathrm{diag}(d_i)$ with $d_i>0$, $G=\mathrm{diag}(g_i)$ with $g_i\ge0$, and let $\eta_i=1$ when row $i$ of $P$ is active and $0$ when it is the zero row. Because the row sums of $J=-D+GP$ are $r_i=\eta_i g_i-d_i$ and $J$ is Metzler, $\min_i r_i\le\alpha(J)\le\max_i r_i$. Equivalently, in stability-margin notation, $\min_i(d_i-\eta_i g_i)\le\mathrm{St}\le\max_i(d_i-\eta_i g_i)$. Likewise $D^{-1}GP$ is nonnegative with row sums $\eta_i g_i/d_i$, so $\min_i(\eta_i g_i/d_i)\le\chi\le\max_i(\eta_i g_i/d_i)$. An isolated actor therefore contributes local margin $d_i$ rather than $g_i-d_i$.
\end{proposition}

\renewcommand{\theproposition}{1B}
\begin{proposition}[Incoming-load envelope]
\label{prop:1B}
Define the gain-weighted incoming load $a_j^{(g)}=\sum_i g_i P_{ij}$ and column sums $c_j=-d_j+a_j^{(g)}$. Applying the same Metzler row-sum bound to $J^\top$ gives $\max\{\min_i r_i,\min_j c_j\}\le\alpha(J)\le\min\{\max_i r_i,\max_j c_j\}$. Equivalently, with row margins $m_i=d_i-\eta_i g_i$ and incoming margins $q_j=d_j-a_j^{(g)}$, $\max\{\min_i m_i,\min_j q_j\}\le\mathrm{St}\le\min\{\max_i m_i,\max_j q_j\}$. The column term is explicitly topology-dependent. When gains are homogeneous, $a_j^{(g)}=g\sum_i P_{ij}$ and is proportional to the attention-share field of Section~\ref{sec:fields}. This is an envelope: column concentration alone need not move the eigenvalue; the actual first-order movement under a feasible rerouting is given by Proposition~\ref{prop:sensitivity} / Remark~\ref{rem:normalized-sensitivity}.
\end{proposition}

\setcounter{proposition}{1}\renewcommand{\theproposition}{\arabic{proposition}}

\textbf{Interpretation.} These bounds are a design consequence of separating routing from gain. $P$ redistributes an actor's coupling across sources but does not create absolute gain. Hence a uniformly subcritical active subsystem stays stable under admissible routing, while the mixed/heterogeneous regime is where topology can assemble different local margins into a collective mode. Proposition~\ref{prop:1B} adds the complementary incoming view: topology can concentrate gain-weighted incoming load on particular sources, tightening a topology-dependent spectral envelope. Proposition~\ref{prop:sensitivity} then identifies which specific feasible reroutings actually move the leading mode.

\begin{definition}[Reactivity]
$$\omega(t) := \lambda_{\max}\!\left(\frac{J(t)+J(t)^\top}{2}\right).$$
$\omega(t)$ bounds the maximal instantaneous growth rate of $\|s(t)\|_2$ and can be positive even when $J(t)$ is Hurwitz, i.e. a system can be asymptotically stable yet exhibit transient amplification.
\end{definition}

\begin{definition}[Finite-horizon transient gain]
$$G_H(t) := \max_{0 \le \tau \le H} \left\| \exp(J(t)\tau) \right\|_2$$
for a fixed horizon $H$, quantifying the largest transient amplification reachable within time $H$.
\end{definition}

\subsection{Critical modes and sensitivity}
\label{sec:critical-modes}

Assume $\lambda_1(t)$, the eigenvalue realizing $\mathrm{St}(t)$, is simple. Let $r_1, l_1$ be its right and left eigenvectors,
$$
J r_1 = \lambda_1 r_1, \qquad l_1^{H} J = \lambda_1 l_1^{H}, \qquad l_1^{H} r_1 = 1.
$$

\begin{proposition}[First-order sensitivity of the leading eigenvalue]
\label{prop:sensitivity}
For $\lambda_1$ simple and isolated, $J = -D+GP$, and $r_1,l_1$ normalized by $l_1^H r_1 = 1$ as above,
$$
\frac{\partial \lambda_1}{\partial P_{ij}} = g_i \cdot \overline{l_{1,i}} \cdot r_{1,j}.
$$
(For the Metzler Jacobian $\lambda_1$ and its eigenvectors are real (Remark~\ref{rem:perron}) and the formula reads $g_i\,l_{1,i}\,r_{1,j}$; the conjugate form covers a complex simple eigenvalue in the signed extension of Remark~\ref{rem:signed}.)
\end{proposition}

This is the standard first-order eigenvalue perturbation formula for a simple eigenvalue, specialized to $J$'s dependence on $P$ through the fixed gain $g_i$; it identifies, for a given actor $i$, which topology edges $P_{ij}$ most move the system's stability margin, and can be used to rank candidate topology perturbations by their first-order effect on that margin (Section~\ref{sec:operational}).

\begin{remark}[Normalized topology perturbations]
\label{rem:normalized-sensitivity}
Proposition~\ref{prop:sensitivity} gives the \emph{ambient} partial derivative with respect to $P_{ij}$, holding the other entries of $P$ fixed. Because $P$ is row-normalized (Definition~\ref{def:P}), its entries are not independent parameters: a feasible perturbation must either preserve the row constraint $\sum_k P_{ik}=1$, or be applied in the evidence space and propagated through the normalization map. For the latter, with $S_i := \sum_k K_{ik} > 0$, the chain rule through $P_{ik}=K_{ik}/S_i$ gives $\partial P_{ik}/\partial K_{ij} = (\delta_{jk}-P_{ik})/S_i$, hence
$$
\frac{\partial\lambda_1}{\partial K_{ij}} = \frac{g_i\,\overline{l_{1,i}}}{S_i}\left(r_{1,j} - \sum_k P_{ik}\,r_{1,k}\right).
$$
The bracketed factor is the deviation of $r_1$ at the target $j$ from actor $i$'s current topology-weighted average of $r_1$: strengthening an edge helps only insofar as it redirects $i$'s coupling toward a part of the critical mode it is not already exposed to, and an edge pointing at the row's current average has zero first-order effect. Intervention sensitivity in the observable evidence space should therefore use this normalized derivative rather than varying a single $P_{ij}$ in isolation.
\end{remark}

\begin{definition}[Critical-mode participation ratio]
$$
PR_{mode}(t) := \frac{\left(\sum_i |r_{1,i}|^2\right)^2}{n \sum_i |r_{1,i}|^4} \in (0,1].
$$
$PR_{mode}$ close to $1$ indicates a broadly distributed critical mode; close to $1/n$, a localized one.
\end{definition}

\begin{remark}[Homogeneous active-subsystem limit as a structural invariance]
\label{rem:homogeneous}
On an irreducible active subsystem with no zero rows, $D=dI$ and $G=gI$ give $J=-dI+gP$ and $\mathrm{St}=d-g$ for every row-stochastic $P$. This is the equality case of Proposition~\ref{prop:1A}: when every interacting actor has the same local margin, conservative routing cannot change the leading stability margin. If the Perron root is simple, the critical mode is the consensus direction $\mathbf 1$. Topology still changes non-leading modes, mixing, transient response and group structure. Once $d_i$ or $g_i$ vary across actors, the invariance breaks; isolated actors remain decoupled with margin $d_i$.
\end{remark}

\subsection{Provenance of the results in this section}
\label{sec:dyn-epistemic}

Propositions~\ref{prop:chi-stability} and~\ref{prop:sensitivity} are standard results of linear systems theory (Metzler/positive-system stability and first-order eigenvalue perturbation), restated in this paper's notation. They apply to $J$ because Section~\ref{sec:state-eq}'s construction makes $J$ the Jacobian of an explicit linear stochastic system, so its hypotheses hold by construction. The theorem of the next section is built on $\mathrm{St}$ as defined here.

% =====================================================================
\section{Model Theorem: From Critical Collective Modes to Measurable Signals}
\label{sec:theorem}
% =====================================================================

A HumanSystem with a Jacobian has a geometry near loss of local stability. That variance rises near a critical transition is classical. The question for BEHAVE is what an observer of this particular relational system should see, and which measurement can miss the transition entirely.

Section~\ref{sec:dynamics} gives the HumanSystem a local Jacobian $J$. This section connects the spectral geometry of $J$ to measurement. As the stability margin $\mathrm{St}$ approaches zero, the slowest collective mode becomes long-lived and stationary fluctuations concentrate along that mode. For a quadratic observable $T(x)=x^\top Qx$, Theorem~\ref{thm:divergence} identifies the leading coefficient $C$ in the divergence and makes it explicit in terms of the critical left/right modes, the diffusion and $Q$.

This creates the bridge that matters for the architecture:
\begin{center}
HumanSystem interaction structure $\to$ local dynamics $J$ $\to$ critical collective mode $\to$ measurement visibility.
\end{center}
The practical content is therefore mode-specific. The theory identifies which collective direction is becoming fragile, how strongly a chosen observable couples to it, and when an otherwise sensible observable is mathematically blind to the critical direction. That is the role of the theorem in BEHAVE.

\subsection{Setup}
\label{sec:thm-setup}

Consider the local dynamics of Section~\ref{sec:dynamics} linearized near an equilibrium of interest and driven by additive noise:
$$
dx = J x\,dt + B\,dW_t, \qquad x(t) \in \mathbb{R}^n,
$$
with $J = -D+GP$ as in Section~\ref{sec:dynamics}, $W_t$ a standard Wiener process, and $\Sigma := BB^\top \succ 0$ the resulting diffusion matrix. Let
$$
T(x) := x^\top Q x, \qquad Q \succeq 0,
$$
be a quadratic observable of the state. (Section~\ref{sec:fields} instantiates $Q$ for specific fields, e.g. the tension field, via a quadratic-form characterization; the theorem below is stated for general $Q \succeq 0$ and does not depend on that instantiation.) Recall $\mathrm{St} := -\max_i \mathrm{Re}\,\lambda_i(J)$ from Definition~\ref{def:stability-margin}; $J$ is Hurwitz for $\mathrm{St}>0$, and $\mathrm{St}\to 0^+$ is the approach to a local stability boundary.

\subsection{An exact identity}
\label{sec:thm-exact}

In the stationary regime (for fixed Hurwitz $J$), the state covariance $\Sigma_x$ solves the continuous Lyapunov equation
$$
J\Sigma_x + \Sigma_x J^\top + \Sigma = 0,
$$
and, exactly, with no approximation,
$$
\mathbb{E}[T] = \mathrm{Tr}(Q\,\Sigma_x).
$$
This holds for every Hurwitz $J$; it is an identity, not yet a statement about behavior near the stability boundary. The theorem below characterizes what happens to $\mathbb{E}[T]$ as $\mathrm{St}\to 0^+$.

\subsection{Theorem}
\label{sec:thm-statement}

Because the statement below is an asymptotic one, $\mathrm{St}\to0^+$, it is a statement about a \emph{family} of Jacobians, not about a single matrix: $\mathrm{St}$ can only change if $J$ does. We therefore state the assumption at the level of that family, $\{J(\varepsilon)\}_{\varepsilon>0}$ with $\varepsilon := \mathrm{St}$, which is what makes the constant $C$ below a genuine constant rather than a quantity that itself drifts with $\mathrm{St}$.

\begin{assumption}[Convergent critical eigendata, with a uniformly stable complement]
\label{ass:spectral}
Let $\{J(\varepsilon)\}_{\varepsilon>0}$ be a family of Jacobians with $\mathrm{St}(J(\varepsilon))=\varepsilon$, and write $\lambda_1(\varepsilon)=-\varepsilon$ for the leading eigenvalue, $r_1(\varepsilon),l_1(\varepsilon)$ for its right and left eigenvectors normalized by $l_1^\top r_1=1$, $\Pi_1(\varepsilon):=r_1 l_1^\top$ for the corresponding spectral projector, and $\Pi_1^\perp := I-\Pi_1$. We assume:
\begin{enumerate}
\item[(A1)] \textbf{Real, simple, gap-separated leading eigenvalue.} For each $\varepsilon>0$, $\lambda_1(\varepsilon)$ is real and algebraically simple (no Jordan block), and is separated from the rest of the spectrum by a fixed gap $\gamma>0$ uniform in $\varepsilon$: $\mathrm{Re}\,\lambda_j(\varepsilon)\le-\varepsilon-\gamma$ for all $j\ne1$.
\item[(A2)] \textbf{Convergence of the family and of the critical eigendata.} $J(\varepsilon)\to J_0$ as $\varepsilon\to0^+$, where $J_0$ has a simple zero eigenvalue with right and left eigenvectors $r_0,l_0$ ($l_0^\top r_0=1$), and the critical spectral projector converges, $\Pi_1(\varepsilon)\to\Pi_0:=r_0l_0^\top$. Consequently $r_1(\varepsilon)\to r_0$ and $l_1(\varepsilon)\to l_0$.
\item[(A2$'$)] \textbf{Rate (for the sharper form only).} $\|J(\varepsilon)-J_0\|_2 = O(\varepsilon)$. Since $\lambda_1(\varepsilon)$ is simple and gap-separated by (A1), standard perturbation theory for a simple eigenvalue then gives $\|\Pi_1(\varepsilon)-\Pi_0\|_2=O(\varepsilon)$ as well (Kato, 1995), which is what upgrades the convergence statement in Theorem~\ref{thm:divergence} to the $O(1)$-remainder form. (A2$'$) holds automatically for any family that is Lipschitz or analytic in $\varepsilon$ at $\varepsilon=0$, including the frozen family used in Section~\ref{sec:thm-numerics}; it is stated separately because it is not implied by (A2) alone.
\item[(A3)] \textbf{Uniformly exponentially stable complement.} There exist constants $M_s,\gamma_s>0$, independent of $\varepsilon$, such that
$$
\left\| \Pi_1^\perp e^{J(\varepsilon)t} \right\|_2 \le M_s\,e^{-\gamma_s t} \qquad \text{for all } t\ge0,\ \varepsilon>0.
$$
\end{enumerate}
\end{assumption}

Two of these three conditions are stronger than a naive reading of ``simple eigenvalue with a spectral gap'' would suggest, and both are necessary for the stated result. (A3) is stronger than the eigenvalue gap alone: for a general non-normal $J$, eigenvalues staying uniformly separated from $\lambda_1$ does not by itself bound how the complementary eigenbasis's conditioning behaves as $\varepsilon\to0^+$, and an arbitrarily ill-conditioned complement could let its contribution to the stationary covariance grow without bound even while every non-critical eigenvalue stays bounded away from $0$; the uniform semigroup bound rules this out at exactly the level the proof uses. (A2) is what makes $C$ well defined: without convergence of $r_1(\varepsilon),l_1(\varepsilon)$, the quantity $(l_1^\top\Sigma l_1)(r_1^\top Qr_1)/2$ would itself be a function of $\varepsilon$, and ``$\mathbb{E}[T]=C/\mathrm{St}+O(1)$ with constant $C$'' would not be a well-posed asymptotic statement. (For the HumanSystem Jacobian $\lambda_1$ is real by Remark~\ref{rem:perron}; realness is stated in (A1) so that the theorem also covers the signed extension of Remark~\ref{rem:signed}, where a leading complex pair is possible and lies outside its scope.)

\begin{theorem}[Divergence of a quadratic observable near the stability boundary]
\label{thm:divergence}
Let $r_0,l_0$ be the limiting critical right and left eigenvectors of (A2), and set
$$
C := \frac{(l_0^\top \Sigma\, l_0)(r_0^\top Q\, r_0)}{2},
$$
a fixed constant determined by the limiting critical mode, the diffusion $\Sigma$, and the observable $Q$. Then:
\begin{enumerate}
\item[(i)] Under (A1)--(A3), $\ \mathrm{St}\cdot\mathbb{E}[T] \to C$ as $\mathrm{St}\to0^+$.
\item[(ii)] If in addition (A2$'$) holds, then
$$
\mathbb{E}[T] = \frac{C}{\mathrm{St}} + O(1) \qquad \text{as } \mathrm{St}\to 0^+,
$$
with no further condition on the diffusion: the critical and complementary subspaces are permitted to mix, $\Pi_1\Sigma(\Pi_1^\perp)^\top \ne 0$.
\end{enumerate}
\end{theorem}

\begin{proof}[Proof (semigroup decomposition)]
Since $J$ is Hurwitz, the stationary covariance has the exact integral representation
$$
\Sigma_x = \int_0^\infty e^{Jt}\,\Sigma\,e^{J^\top t}\,dt,
$$
and $\mathbb{E}[T] = \mathrm{Tr}(Q\Sigma_x)$ (Section~\ref{sec:thm-exact}). Split $e^{Jt} = \Pi_1 e^{Jt} + \Pi_1^\perp e^{Jt} = e^{\lambda_1 t}\Pi_1 + \Pi_1^\perp e^{Jt}$, using $\lambda_1$ real and $\Pi_1 J = J\Pi_1 = \lambda_1\Pi_1$. Substituting into $\Sigma_x$ gives three terms: a critical-critical term, a critical-complement cross term (and its transpose), and a complement-complement term.

\textbf{Critical-critical term.} $\int_0^\infty e^{2\lambda_1 t}\,dt\cdot \Pi_1\Sigma\Pi_1^\top = \frac{1}{2\varepsilon}\,\Pi_1\Sigma\Pi_1^\top$, since $\lambda_1=-\varepsilon$. Projected through $Q$, its trace contribution is $\frac{1}{2\varepsilon}\,\mathrm{Tr}(Q\,\Pi_1\Sigma\Pi_1^\top) = \frac{C(\varepsilon)}{\varepsilon}$ with $C(\varepsilon):=(l_1(\varepsilon)^\top\Sigma\,l_1(\varepsilon))(r_1(\varepsilon)^\top Q\,r_1(\varepsilon))/2$, using $\Pi_1=r_1l_1^\top$ and $l_1^\top r_1=1$. By (A2), $r_1(\varepsilon)\to r_0$ and $l_1(\varepsilon)\to l_0$, hence $C(\varepsilon)\to C$. Under (A2$'$) we have in addition $\|\Pi_1(\varepsilon)-\Pi_0\|_2=O(\varepsilon)$, and since $C(\varepsilon)=\mathrm{Tr}(Q\,\Pi_1\Sigma\Pi_1^\top)/2$ is a smooth (quadratic) function of $\Pi_1$ on a bounded set, $C(\varepsilon)=C+O(\varepsilon)$.

\textbf{Complement-complement term.} By the uniform bound in Assumption~\ref{ass:spectral}, $\|\Pi_1^\perp e^{Jt}\|_2 \le M_s e^{-\gamma_s t}$ with $M_s,\gamma_s$ independent of $\mathrm{St}$, so
$$
\left\| \int_0^\infty \Pi_1^\perp e^{Jt}\,\Sigma\,e^{J^\top t}(\Pi_1^\perp)^\top\,dt \right\|_2 \le \|\Sigma\|_2 \int_0^\infty M_s^2 e^{-2\gamma_s t}\,dt = \frac{M_s^2\|\Sigma\|_2}{2\gamma_s},
$$
a constant independent of $\mathrm{St}$; its trace contribution through $Q$ is therefore $O(1)$.

\textbf{Cross terms.} These are bounded directly, without Cauchy--Schwarz. The critical factor contributes $e^{\lambda_1 t}=e^{-\varepsilon t}$ and the complementary factor is controlled by (A3), so their product decays at rate $\varepsilon+\gamma_s$, which is bounded below by $\gamma_s$ \emph{uniformly in} $\varepsilon$:
$$
\left\| \int_0^\infty e^{-\varepsilon t}\,\Pi_1\Sigma\,(\Pi_1^\perp e^{Jt})^\top dt \right\|_2 \le \|\Pi_1\|_2\|\Sigma\|_2 M_s \int_0^\infty e^{-(\varepsilon+\gamma_s)t}\,dt = \frac{\|\Pi_1\|_2\|\Sigma\|_2 M_s}{\varepsilon+\gamma_s} \le \frac{\|\Pi_1\|_2\|\Sigma\|_2 M_s}{\gamma_s},
$$
and $\|\Pi_1(\varepsilon)\|_2$ is itself uniformly bounded, since $\Pi_1(\varepsilon)\to\Pi_0$ by (A2). The transposed cross term is bounded identically. Both are therefore $O(1)$, with no assumption about whether the critical and complementary subspaces mix through $\Sigma$: the slowly decaying critical factor multiplies a uniformly fast-decaying complementary factor, and it is the faster rate that controls the product.

\textbf{Combining.} Summing the three contributions,
$$
\mathbb{E}[T] = \frac{C(\varepsilon)}{\varepsilon} + O(1),
$$
where the $O(1)$ term collects the complement-complement and both cross terms, and is bounded as $\varepsilon\to0^+$ by the constants $M_s,\gamma_s,\|\Sigma\|_2,\|Q\|_2$ and $\sup_\varepsilon\|\Pi_1(\varepsilon)\|_2$ alone, not by any property of $J(\varepsilon)$ that could itself degrade as $\varepsilon\to0$, which is exactly what (A2) and (A3) are designed to rule out. Multiplying by $\varepsilon$ and using $C(\varepsilon)\to C$ gives claim~(i), $\varepsilon\,\mathbb{E}[T]\to C$. Under (A2$'$), $C(\varepsilon)=C+O(\varepsilon)$, so $C(\varepsilon)/\varepsilon = C/\varepsilon + O(1)$, giving claim~(ii), $\mathbb{E}[T]=C/\mathrm{St}+O(1)$.
\end{proof}

\begin{remark}[Cross-term order]
The cross terms in claim~(ii) are $O(1)$ because the two factors decay \emph{simultaneously}, at combined rate $\varepsilon+\gamma_s\ge\gamma_s$, so no condition on the diffusion is needed. (A Cauchy--Schwarz bound would give only $O(1/\sqrt{\mathrm{St}})$ unless $\Pi_1\Sigma(\Pi_1^\perp)^\top=0$.) The direct estimate above is uniform in $\varepsilon$, so claim~(ii) holds for a general non-normal family whose critical and complementary subspaces do mix through $\Sigma$, not only for the normal case. Section~\ref{sec:thm-numerics} reports a numerical check of exactly this: a non-normal family with $\|\Pi_1\Sigma(\Pi_1^\perp)^\top\|_2\approx1.6$, whose remainder is observed to converge to a constant rather than to grow like $\mathrm{St}^{-1/2}$.
\end{remark}

\begin{remark}
Let $z(t) := l_1^\top x(t)$ be the leading-mode coordinate; since $l_1^\top J=\lambda_1 l_1^\top$, $z$ is itself a scalar Ornstein--Uhlenbeck process, $dz=-\mathrm{St}\cdot z\,dt+\sigma_1\,dW_t$ with $\sigma_1^2:=l_1^\top\Sigma\,l_1$, stationary variance $\mathrm{Var}(z)=\sigma_1^2/(2\,\mathrm{St})$, and $\mathbb{E}[z]\equiv0$ for every $\mathrm{St}>0$. Only the second moment of the leading mode diverges: a linear observable $\ell(x)=l_1^\top x$ does not diverge as $\mathrm{St}\to0^+$, even though its variance does. The divergence in Theorem~\ref{thm:divergence} is specific to quadratic (or higher even-order) observables; it is a consequence of critical slowing down, $\tau_1 = 1/\mathrm{St}\to\infty$, acting on the leading mode's variance, not its mean.
\end{remark}

\begin{remark}[Relation to prior internal work]
The $1/\mathrm{St}$ scaling was first noticed in an internal technical record predating this paper, as a diagnostic observation without spectral assumptions or proof. Turning that observation into Theorem~\ref{thm:divergence} required identifying exactly which conditions it depends on (Assumption~\ref{ass:spectral}), proving it under them, and establishing where it fails (Section~\ref{sec:thm-numerics}).
\end{remark}

\begin{corollary}[Divergence is observable-dependent]
\label{cor:C-positive}
In Theorem~\ref{thm:divergence}, $C\ge0$ always, since $\Sigma\succ0$ and $Q\succeq0$. If $r_0^\top Q r_0 > 0$, i.e. the quadratic observable has nonzero projection onto the limiting critical mode, then $C>0$ and $\mathbb{E}[T]$ diverges as $\mathrm{St}\to0^+$. If instead $r_0^\top Q r_0 = 0$, i.e. the observable is exactly orthogonal to that mode, then $C=0$, claim~(i) degenerates to $\mathrm{St}\cdot\mathbb{E}[T]\to0$, and this particular observable does not register the approaching instability at the $1/\mathrm{St}$ rate, even though the underlying dynamics still has $\mathrm{St}\to0^+$: not every computable quadratic quantity is a witness to every critical mode, and which one is chosen matters.
\end{corollary}

\textbf{Example (A quadratic discord numerator that is exactly blind in the homogeneous limit).} In the irreducible homogeneous case of Remark~\ref{rem:homogeneous}, the critical right mode is the consensus vector $\mathbf 1$. The quadratic numerator $N_L=\mathrm{tr}(Z_c^\top L_G Z_c)$ of graph discord is Laplacian-based and annihilates the consensus direction, so its critical projection is zero and it does not acquire the $C/\mathrm{St}$ divergence from that mode. The normalized field $L$ itself is a ratio, not the quadratic observable of Theorem~\ref{thm:divergence}. Section~\ref{sec:fields} therefore actor-centers $Z$ at each instant and interprets $L$ as disagreement geometry after removal of the common mode. It is a concrete instance of why multiple observables are required.

\begin{corollary}[Application to the quadratic kinematic component of tension]
\label{cor:tension}
Let $Q_T$ denote only the fixed-baseline quadratic kinematic component used inside the operational tension field, not the full operational tension score with its proxemic term or adaptive normalization. Under Assumption~\ref{ass:spectral}, and provided the critical mode has nonzero projection onto these coordinates, this component inherits the leading $C_T/\mathrm{St}$ law. This corollary is therefore an example of how a declared quadratic measurement couples to the critical mode; it is not a theorem that the complete online tension implementation is exactly $x^\top Q_T x$ in the same coordinates over every deployment window.
\end{corollary}

\subsection{Numerical verification}
\label{sec:thm-numerics}

Verification uses a single frozen family of Jacobians: a fixed eigenbasis and fixed non-critical eigenvalues, with only $\lambda_1=-\mathrm{St}$ varied across runs, so that the trend as $\mathrm{St}\to 0$ is not confounded with resampling noise from rebuilding $J$ at each point. $Q=I$, $\Sigma=I$, so Theorem~\ref{thm:divergence} predicts $C=0.5$ exactly.

\begin{table}[H]
\centering\small
\begin{tabular}{@{}rrr@{}}
\toprule
$\mathrm{St}$ & $\mathbb{E}[T]$ (exact, Lyapunov) & $\mathrm{St}\cdot\mathbb{E}[T]$ \\
\midrule
$0.2$        & $3.9162$    & $0.7832$ \\
$0.1$        & $6.4162$    & $0.6416$ \\
$0.05$       & $11.4162$   & $0.5708$ \\
$0.0125$     & $41.4162$   & $0.5177$ \\
$0.003125$   & $161.4162$  & $0.5044$ \\
$0.00078125$ & $641.4162$  & $0.5011$ \\
$0.000390625$& $1281.4162$ & $0.5006$ \\
\bottomrule
\end{tabular}
\caption{Frozen family ($n=5$, $Q=I$, $\Sigma=I$; single fixed eigenbasis, only $\lambda_1=-\mathrm{St}$ varied): $\mathrm{St}\cdot\mathbb{E}[T]\to 0.5006$, against the analytic prediction $C=0.5$. Tail-fit exponent (last 5 points) $0.9943$ against a theoretical exponent of $1$, $R^2=0.999995$. A separate \emph{pointwise} robustness check (which resamples a fresh random symmetric matrix with controlled $\lambda_1$ at each $\mathrm{St}$ and is therefore not an instance of the convergent-family assumptions of Theorem~\ref{thm:divergence}) converges to the same constant (last $\mathrm{St}\,\mathbb{E}[T]=0.5005$, tail exponent $0.9952$, $R^2=0.999997$). It is reported as a sanity check rather than as a test of the theorem's hypotheses: for $Q=\Sigma=I$ and a normalized real symmetric critical mode, $C=1/2$ regardless of that mode's orientation, so the check is meaningful without requiring the resampled matrices to form a convergent family. A fixed-step Euler--Maruyama discretization covariance check on the frozen family agrees with the continuous Lyapunov solution to $0.32\%$ at $\mathrm{St}=0.2$ and $0.11\%$ at $\mathrm{St}=0.05$.}
\end{table}

\textbf{Non-normal check (remainder order with cross-diffusion).} Claim~(ii) asserts an $O(1)$ remainder with no condition on whether the critical and complementary subspaces mix through $\Sigma$. To check this where it is not automatic, we use a family with a fixed \emph{non-orthogonal} eigenbasis ($n=4$, $Q=\Sigma=I$, non-critical eigenvalues fixed at $-1.5,-2.5,-3.5$, only $\lambda_1=-\mathrm{St}$ varied), for which $\|\Pi_1\Sigma(\Pi_1^\perp)^\top\|_2\approx1.62$, i.e. the cross-diffusion term is substantial. The observed remainder $\mathbb{E}[T]-C/\mathrm{St}$, with $C=1.0969$ computed from the limiting critical eigenvectors, is $-0.326$, $-0.469$, $-0.509$, $-0.520$, $-0.522$, $-0.523$ at $\mathrm{St}=0.2$ down to $1.95\times10^{-4}$: it converges to a constant rather than growing, and $(\mathbb{E}[T]-C/\mathrm{St})\sqrt{\mathrm{St}}\to0$, showing that the remainder does not grow or scale as $\mathrm{St}^{-1/2}$ for this non-normal, cross-diffusing family.

\textbf{Negative control (degenerate leading eigenvalue).} Replacing Assumption~\ref{ass:spectral} with a $2\times2$ Jordan block (defective $\lambda_1$, algebraic multiplicity $2$) changes the scaling: the fitted exponent is $\approx 2.838$ against a theoretical exponent of exactly $3$ (i.e. $\mathbb{E}[T]\sim \mathrm{St}^{-3}$, not $\mathrm{St}^{-1}$) for the defective case. This shows the simplicity assumption in Assumption~\ref{ass:spectral} is essential for the stated scaling: violating it changes the divergence rate itself. For the HumanSystem Jacobian a defective leading eigenvalue cannot arise when the coupling is irreducible (Remark~\ref{rem:perron}); it requires one-way-connected subgroups with equal Perron roots. The Jordan block is therefore a mathematical negative control for Assumption~\ref{ass:spectral}, not the expected failure mode of an irreducible HumanSystem.

\textbf{Reducible topology: a sanity check outside Assumption~\ref{ass:spectral}.} If the population splits into subgroups with no interaction evidence between them, $P$ is block-diagonal and each block contributes its own Perron eigenvalue; with homogeneous $D,G$ these coincide, giving a repeated but semisimple leading eigenvalue: a zero spectral gap without a Jordan block, which is the natural structural degeneracy of a HumanSystem. This violates the simplicity condition of (A1). A two-block check (two $3$-actor blocks with independent random row-stochastic topologies, $D=I$, $G=(1-\mathrm{St})I$; geometric multiplicity of the leading eigenvalue confirmed to be $2$) nonetheless exhibits the same $1/\mathrm{St}$ exponent: tail-fit $0.995$ ($R^2=0.999997$), against $3$ for the defective case. This suggests that a multidimensional semisimple extension of Theorem~\ref{thm:divergence} may be possible; that extension is not proved here.

\begin{figure}[H]
\centering
\includegraphics[width=0.72\textwidth]{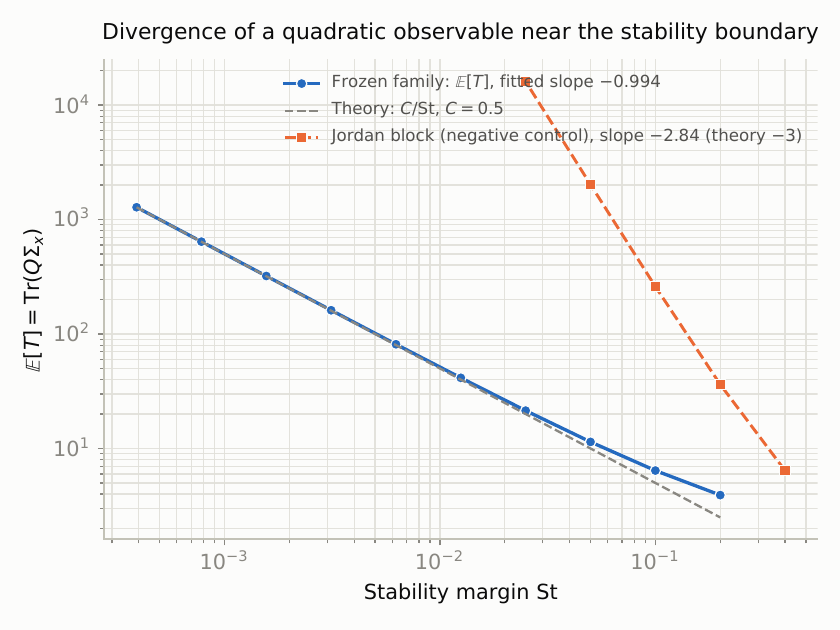}
\caption{Exact computation for the frozen family: $\mathbb{E}[T]=\mathrm{Tr}(Q\Sigma_x)$ solved exactly from the Lyapunov equation at each $\mathrm{St}$, log-log against $\mathrm{St}$. The fitted slope ($-0.994$) matches the theorem's predicted exponent of $-1$; the dashed line is the analytic asymptote $C/\mathrm{St}$ with $C=0.5$. The Jordan-block negative control (orange, dash-dot), computed the same way but with a defective leading eigenvalue, has a visibly different slope ($-2.84$, theory $-3$).}
\label{fig:theorem-convergence}
\end{figure}

\begin{figure}[H]
\centering
\includegraphics[width=\textwidth]{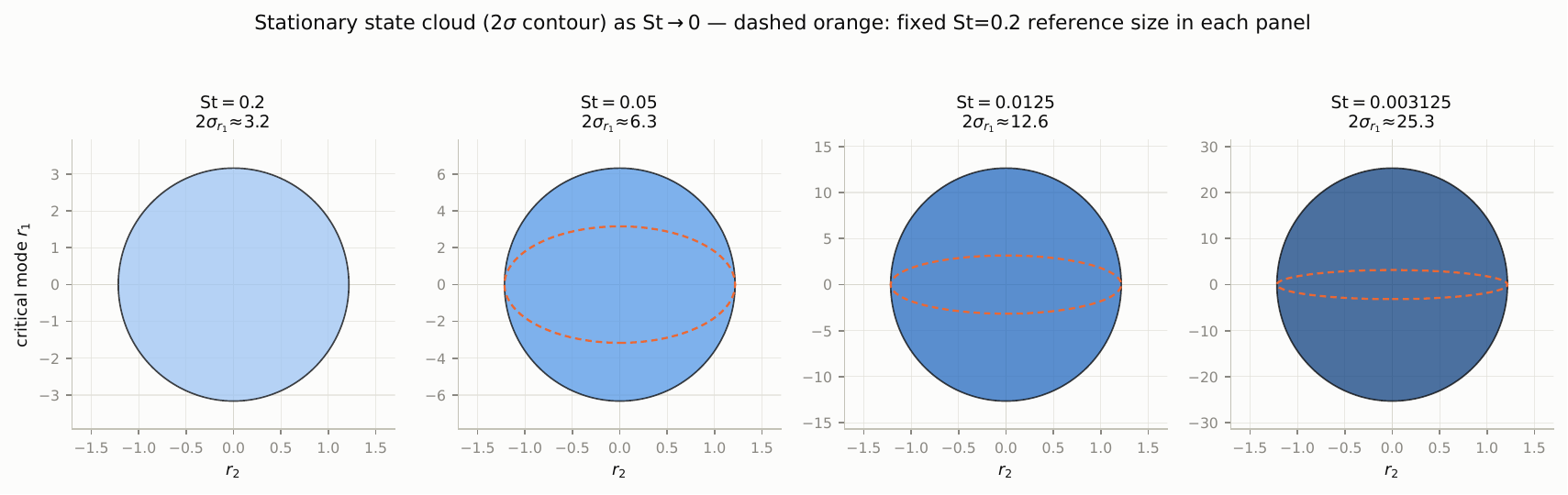}
\caption{The geometry of fragility. Stationary $2\sigma$ covariance contour of the state for the frozen family of Section~\ref{sec:thm-numerics}, projected onto the plane of the critical mode $r_1$ and a fixed non-critical mode $r_2$ (orthonormal eigenvectors of $J$, so the covariance is diagonal in this basis and each axis is read directly from $r_k^\top\Sigma_x r_k$). The dashed orange outline repeats the $\mathrm{St}=0.2$ ellipse at fixed size in every panel. As $\mathrm{St}\to0$ the $r_2$ extent stays put while the $r_1$ extent grows $\propto\mathrm{St}^{-1/2}$ ($3.2\to6.3\to12.6\to25.3$, doubling for every $4\times$ drop in $\mathrm{St}$, as $\mathrm{Var}=\sigma_1^2/(2\,\mathrm{St})$ predicts). The system does not simply become more variable: its uncertainty stretches along one specific, computable collective direction. This is the picture behind ``distance to the stability boundary'': a spectral quantity of the dynamics rather than a Euclidean distance between raw measurements.}
\label{fig:theorem-geometry}
\end{figure}

\subsection{Scope}
\label{sec:thm-limits}

\begin{enumerate}
\item \textbf{Linearization.} The theorem concerns dynamics linearized near an equilibrium, and makes no claim about behavior far from equilibrium in a nonlinear regime.
\item \textbf{Noise model.} Additive Gaussian noise with state-independent $\Sigma$ is assumed; multiplicative or non-Gaussian noise would require a separate derivation.
\item \textbf{Spectral assumption.} A defective (Jordan) leading eigenvalue violates Assumption~\ref{ass:spectral}; Section~\ref{sec:thm-numerics} shows it changes the exponent from $1$ to $3$. A leading complex pair cannot occur for the Metzler HumanSystem Jacobian (Remark~\ref{rem:perron}) and arises only in the signed extension. A repeated but semisimple leading eigenvalue, produced by reducible topology, lies outside (A1) as stated; the sanity check of Section~\ref{sec:thm-numerics} finds the same exponent there, and a semisimple extension is open.
\end{enumerate}

Theorem~\ref{thm:divergence} applies to the local stochastic model under Assumption~\ref{ass:spectral}; estimating its spectral quantities from finite trajectory windows uses the parameter identification of Section~\ref{sec:identification}.

% =====================================================================
\section{Derived Fields and HumanSystemState}
\label{sec:fields}
% =====================================================================

The fields defined in this section are \emph{system-referenced derived observables}: quantities whose values are defined only relative to the system they are computed in. They come in two kinds.

\textbf{Intrinsically relational fields} cannot be written as a function of any single actor's trajectory at all, because their definition quantifies over the interaction structure: topology-aware discord $L$ (a Rayleigh quotient over the interaction graph), attention share $A_j$ (a normalized column of $P$), direct reach $I_{direct}$ and critical leverage $I_{crit}$ (the latter a function of the Jacobian's left critical eigenvector), and the participation ratio $PR_T$ (a distributional property of the population). For these, a complete per-actor description does not determine the field value; the interaction structure is an irreducible argument.

\textbf{System-normalized actor-level fields} are computed from per-actor kinematics but are referenced to the system: tension $T_i$ and momentum $M_i$ are built from an individual's speed and gestural amplitude, z-scored against a population baseline and, for $T_i$, augmented by a local proxemic term. Here the actor-level signal is the substrate, and the system supplies the normalization and local context that make the number comparable: the same raw kinematics yield different field values in different systems.

Both kinds are reported in $S_{\mathcal H}(t)$ and used downstream. For the first kind, the quantity has no value at the level of a single tracked actor.

\subsection{Feature normalization}
\label{sec:feat-norm}

Fields are computed from simple, declared functions of the observability layer's outputs (Section~\ref{sec:obs-layer}), not from additional hidden inputs. For actor $i$, write $z^v_i := (\|v_i\|-\mu_v)/\sigma_v$ for the z-scored speed and $z^e_i := (\tilde e_i - \mu_e)/\sigma_e$ for a z-scored gestural amplitude $\tilde e_i$ (normalized gestural signal amplitude, when reliably extractable), with $\mu_v,\sigma_v,\mu_e,\sigma_e$ estimated from a trailing baseline window. $p_i$ denotes a declared local proxemic-pressure term (e.g. a function of local density/spacing), in the sense of the proxemic literature originating with Hall (1966). These are the only per-actor quantities the fields below draw on beyond $Y(t)$, $K(t)$, $P(t)$.

\subsection{Fields}
\label{sec:fields-defs}

\begin{lemma}[Characterization of the intensity form]
\label{lem:quadratic-T}
Let $\varphi:\mathbb{R}^2\to\mathbb{R}$ be a polynomial in the two normalized channels $(v,e)$ satisfying:
\begin{enumerate}
\item[(P1)] \textbf{Sign invariance.} $\varphi(\pm v,\pm e)=\varphi(v,e)$: deviation from baseline counts the same in either direction.
\item[(P2)] \textbf{Channel additivity.} $\varphi(v,e)=\varphi_v(v)+\varphi_e(e)$ for univariate polynomials $\varphi_v,\varphi_e$: no cross term between channels.
\item[(P3)] \textbf{Vanishing at baseline.} $\varphi(0,0)=0$.
\item[(P4)] \textbf{Degree-2 homogeneity.} $\varphi(\alpha v,\alpha e)=\alpha^2\varphi(v,e)$ for all $\alpha>0$.
\item[(P5)] \textbf{Monotonicity in magnitude.} $\varphi_v$ is nondecreasing in $|v|$ and $\varphi_e$ nondecreasing in $|e|$.
\end{enumerate}
Then $\varphi(v,e)=\gamma_v v^2+\gamma_e e^2$ with $\gamma_v,\gamma_e\ge0$, and this form is unique up to the choice of those two weights.
\end{lemma}

\begin{proof}
By (P2), $\varphi=\varphi_v+\varphi_e$; by (P3) and additivity we may take $\varphi_v(0)=\varphi_e(0)=0$. Setting $e=0$ in (P4) gives $\varphi_v(\alpha v)=\alpha^2\varphi_v(v)$ for all $\alpha>0$, so $\varphi_v$ is a univariate polynomial homogeneous of degree $2$, i.e. $\varphi_v(v)=\gamma_v v^2$; symmetrically $\varphi_e(e)=\gamma_e e^2$. These are even, so (P1) is satisfied automatically. Finally (P5) forces $\gamma_v,\gamma_e\ge0$.
\end{proof}

\begin{remark}[What this does and does not establish]
The work here is done by (P4): degree-2 homogeneity is what fixes the degree, and once the degree is fixed, additivity and evenness leave no freedom beyond the two nonnegative weights. The lemma is therefore a statement about what the declared invariances permit, not evidence that human behavioral intensity \emph{is} quadratic in normalized kinematics. It is used in one specific way (as the premise fixing the functional form of $T_i$ below, and the premise instantiating $Q$ in Corollary~\ref{cor:tension}) and carries no empirical content of its own. A deployment for which (P4) is the wrong scaling assumption would obtain a different form, and nothing else in this paper would change.
\end{remark}

\begin{definition}[Tension]
$$T_i(t) := \gamma_v\,(z^v_i)^2 + \gamma_e\,(z^e_i)^2 + \gamma_p\,p_i, \qquad \gamma_v,\gamma_e,\gamma_p \ge 0.$$
The quadratic kinematic terms follow Lemma~\ref{lem:quadratic-T}; the proxemic term $\gamma_p p_i$ is an explicit additional modeling choice, linear by construction, not derived from the same axioms. Acceleration does not enter $T_i$.
\end{definition}

\begin{definition}[Momentum]
$$M_i(t) := \left\| \frac{d}{dt} z_i(t) \right\|_2, \qquad M(t) := \mathrm{RMS}_i\,M_i(t),$$
where $z_i(t)$ is the normalized per-actor kinematic state. $M_i$ is a rate of change of state; it carries acceleration-like information and is not a mean-speed proxy.
\end{definition}

\begin{definition}[Topology-aware discord]
Let $A_{sym}=(P+P^\top)/2$, $L_G=D_A-A_{sym}$ its graph Laplacian, and let $C_n=I-(1/n)\mathbf 1\mathbf 1^\top$. For a matrix $Z$ of per-actor state channels at the current instant, define $Z_c=C_nZ$: actor-centering across the current group, not temporal centering. Then
$$L(t)=\frac{\mathrm{tr}(Z_c^\top L_G Z_c)}{\lambda_{\max}(L_G)\,\mathrm{tr}(Z_c^\top Z_c)+\epsilon}\in[0,1].$$
$L$ is a normalized graph Rayleigh quotient: it measures whether interacting neighbours disagree after the common actor mode has been removed. Its quadratic numerator $N_L=\mathrm{tr}(Z_c^\top L_G Z_c)$ is the object that can be compared directly with Corollary~\ref{cor:C-positive}; $L$ itself is a ratio and is not claimed to obey the theorem's $C/\mathrm{St}$ law.
\end{definition}

\begin{definition}[Attention share]
With $P_{ij}$ read as source $j\to$ receiver $i$,
$$A_j(t) := \frac{\sum_i P_{ij}(t)}{\sum_{i,k} P_{ik}(t)}, \qquad \sum_j A_j = 1.$$
Attention concentrated on a source is a column quantity under this indexing convention.
\end{definition}

\begin{definition}[Direct reach and critical leverage]
$$I_{direct,j}(t) := \sum_i (G(t)P(t))_{ij}, \qquad I_{crit,j}(t) := \frac{|l_{1,j}|}{\sum_k |l_{1,k}|}.$$
$I_{direct}$ is instantaneous local reach; $I_{crit}$ is cascade leverage via the critical left mode of Section~\ref{sec:critical-modes}. These are kept as two separate fields: the right-eigenvector centrality of a row-normalized $P$ is not used as a stand-in for either.
\end{definition}

\begin{definition}[Phase coherence and directional coordination]
When a genuine oscillatory signal and a defensible instantaneous phase $\phi_i(t)$ are identifiable (per the observability-first principle, Section~\ref{sec:obs-first}),
$$S_{phase}(t) := \frac{\left|\sum_i w_i e^{i\phi_i}\right|}{\sum_i w_i}.$$
Otherwise, directional (non-oscillatory) coordination is measured separately by
$$C_{dir}(t) := \frac{\left\|\sum_i w_i\, v_i/\|v_i\|\right\|}{\sum_i w_i} \in [0,1].$$
Body orientation is not a Kuramoto phase: $S_{phase}$ is computed only where identifiability is established for the given signal, never substituted with orientation by default.
\end{definition}

\begin{definition}[Residual fluctuation]
With $\xi_i(t) := \dot s_i + d_i s_i - g_i(Ps)_i - b_i^\top u_{ext}(t)$ the model residual from Section~\ref{sec:dynamics},
$$N(t) := \mathrm{tr}\,\mathrm{Cov}_{window}(\xi(t)).$$
$N$ is the unexplained fluctuation level after the local-dynamics model, not a second measure of the variance already captured by $T$ or $L$.
\end{definition}

\begin{definition}[Spatial heterogeneity]
$$B(t,x) := \|\nabla \bar T(t,x)\|.$$
\end{definition}

\begin{definition}[Participation ratio]
For nonnegative per-actor activity $T_1,\dots,T_n$,
$$PR_T(t) := \frac{\left(\sum_i T_i\right)^2}{n\sum_i T_i^2} \in [1/n,\,1].$$
High $T$ with low $PR_T$ indicates activity concentrated in a few actors; high $T$ with high $PR_T$ indicates distributed activation. $T$ and $PR_T$ are reported as separate axes, not collapsed into one score, specifically to avoid a false-positive reading of aggregate intensity.
\end{definition}

\subsection{Field audit}
\label{sec:field-audit}

Table~\ref{tab:field-audit} records where measurement against real trajectory data changed a field's definition, and what forced each change. This is the mechanism by which the field layer was disciplined: a field that failed to carry distinct information was not kept for completeness, it was rebuilt. Each row names the diagnostic that caught the problem, so every change is independently checkable.

\begin{table}[H]
\centering\small
\begin{tabularx}{\textwidth}{@{}p{1.6cm}p{4.7cm}p{4.7cm}p{3.3cm}@{}}
\toprule
\textbf{Field} & \textbf{Earlier definition} & \textbf{Corrected definition} & \textbf{Reason} \\
\midrule
$L$ & Raw per-actor state dispersion & Graph Rayleigh quotient over interacting neighbors (topology-aware discord) & Measured $\mathrm{corr}(T,L)=0.996$ on real data with the raw-dispersion definition (near-redundant with $T$); $-0.04$ after correction \\
$M$ & Windowed aggregate of mean speed & $\|dz_i/dt\|_2$, the rate of change of normalized state & Mean-speed proxy does not carry acceleration; must not be reintroduced as momentum \\
$A$ & Row sum of $P$ called ``group attention'' & Column sum $A_j=\sum_i P_{ij}/\sum_{ik}P_{ik}$ & Index-convention error: attention on a source is a column quantity under the $j\to i$ convention \\
$I$ & Single ``source influence'' from the right eigenvector of row-normalized $P$ & Split into $I_{direct}$ (from $GP$) and $I_{crit}$ (from left critical eigenvector $l_1$) & Conflated instantaneous reach with cascade-trigger leverage, two distinct quantities \\
$S$ & Body-orientation alignment treated as oscillatory phase & $S_{phase}$ restricted to identifiable oscillatory signals; $C_{dir}$ for directional coordination otherwise & Orientation is not a Kuramoto phase; computing $S_{phase}$ regardless of identifiability violates the observability-first principle \\
$T$ & Included an acceleration term & Restricted to the quadratic kinematic form of Lemma~\ref{lem:quadratic-T} plus an explicit linear proxemic term & Acceleration reassigned to $M$; keeps $T$'s quadratic-form derivation clean \\
\bottomrule
\end{tabularx}
\caption{Field audit: corrected field definitions. Each row is independently falsifiable against the stated reason.}
\label{tab:field-audit}
\end{table}

\subsection{HumanSystemState}
\label{sec:SH-def}

\begin{definition}[HumanSystemState]
$$S_{\mathcal H}(t) := F_S\big(Y(t), K(t), P(t), \mathcal{G}(t), \mathcal{F}(t), \mathrm{history}\big),$$
for some declared, computable $F_S$. $S_{\mathcal H}(t)$ may include global, group-specific, and spatial components (e.g. regime, coordination, momentum, spatial structure, fragility, and an observability/confidence status), selected by task. $S_{\mathcal H}(t)$ is not defined as a universal fixed-length vector: which fields and structural summaries it includes is open and task-dependent, consistent with $\mathcal{H}(t) \neq S_{\mathcal H}(t)$ (Definition~\ref{def:H-SH}): the identity of the object does not depend on which components of its state a given task chooses to compute.
\end{definition}

$S_{\mathcal H}(t)$ is the answer to the question a machine operating among people actually needs answered: what is this collective like now? It combines structure (which groups exist, where their boundaries and bridges are), fields (tension, discord, attention, momentum, coordination), dynamics (stability margin, critical mode, fragility) and observability (how well each of these is currently measured) into one state of one object. Changes in it are state transitions; changes in it under an explicit perturbation are responses (Section~\ref{sec:operational}).

% =====================================================================
\section{Empirical Evaluation}
\label{sec:empirical}
% =====================================================================

\textbf{Where this section sits in the architecture.} The experiments below evaluate the architecture from the relational representation layer through a first causal stress test of the operator layer. Representation-level tests ask whether interaction structure carries information absent from conventional descriptors, whether relationally distinct collective states evolve differently, whether the collective responds measurably to an explicit perturbation, and whether a collective state computed causally from trailing observables can change before a downstream operational consequence. Section~\ref{sec:operator-stress} then tests the next layer directly by fitting the structured local operator causally, evaluating held-out dynamics, comparing classical early-warning baselines, and relating the fitted spectral margin to a separately defined future throughput outcome. This remains a stress test rather than full errors-in-variables identification, but the spectral layer is therefore no longer only conditional on future empirical work.

Each result carries an evidence label: \textsc{Validated} (reproducible on public data with an audited, seed-fixed analysis), \textsc{Supported} (reproducible on public data with a public script, statistics as reported by that script), \textsc{Candidate} (positive, but provenance or robustness not yet at that standard), \textsc{Restricted} (from consented recordings that cannot be redistributed), \textsc{Exploratory}, \textsc{Negative boundary}, and \textsc{Open}.

\subsection{DIAMOR: group-structure discrimination}
\label{sec:diamor}

On DIAMOR-1 (1271 annotated pairs, 9279 agents), adding body-facing evidence to proximity for group-membership discrimination raised AUC from $0.8955$ to $0.9331$ ($\Delta=+0.0376$; permutation test, $1000$ permutations, $z=+22.31$, $p<0.001$; seed $42$). DIAMOR-2 (677 pairs, separate recording day) replicated the direction with an increment of $+0.060$ at a smaller permutation $z$-score ($z\approx17$), consistent with its smaller sample. \textsc{Validated}: interaction evidence beyond raw proximity carries group-membership information; effect size differs by dataset.

Within-pair time-series synchrony on DIAMOR-1: direction-of-motion synchrony $+0.270$ vs.\ $+0.023$ baseline ($p=.0002$); body-facing synchrony $+0.201$ vs.\ $-0.129$ ($p=.0001$); $72\%$ side-by-side configuration. Speed synchrony was not significant ($p=.17$).

\textbf{Scope of the validated weight.} The validated result above uses one specific, time-averaged, symmetric weight, $\mathrm{prox}\cdot(1+\cos\Delta_{face})/2$. It does not validate every implementation of $K$. A separate check of a directional bearing kernel (co-presence controls, $h_r=1.4\,\mathrm{m}$) changed AUC by $\approx-0.004$, and by $\approx-0.008$ at a default, per-frame median-distance $h_r$ (which can reach $\approx10\,\mathrm{m}$ in a corridor, blurring proximity); that kernel does carry real directional information (real headings outperform trajectory-permuted headings) but its shape is a poor fit for side-by-side walking partners. \textsc{Negative boundary}: kernel form is part of the empirical model, not guaranteed by the interaction-structure idea itself, and this specific kernel does not reproduce the validated effect.

\subsection{Perturbation probes: walkers passing through a group}
\label{sec:probes}

A passing mobile agent as an explicit, dated perturbation: in a Warsaw event recording, stationary agents' dispersion increased $1.6\times$ (speed $p=.0014$, gesture $p=.0032$). On DIAMOR-1 ($2735$ events), group members suppressed motion (speed-variance ratio $.796$, $p<10^{-16}$; orientation-change ratio $.639$, $p<10^{-27}$). \textsc{Validated} on DIAMOR-1; \textsc{Restricted} for the Warsaw recording (Section~\ref{sec:restricted}). On DIAMOR-2 ($11175$ events), the same direction of effect did not reach significance ($p=.11$), which marks a replication boundary. Orientation did not predict which specific agent reacted better than chance ($p=.11$), and reaction was not distance-dependent ($p=.997$): \textsc{Negative} result for exact-reactor localization, which remains open.

\subsection{Jülich: matched-state divergence}
\label{sec:julich-matched}

The original public script builds local flow states from J\"ulich bottleneck trajectories in $2\,$m cells and matches moments on density, mean speed and flow. For this version we reran the analysis on all 24 uploaded 2021 bottleneck trajectories with speed dispersion added as a fourth standardized matching variable and with the experimental run, not the neighbouring time window, as the unit of inference. Nineteen runs yielded at least five calm-vs-breakdown matched pairs (3639 pairs total). Future-breakdown states retained higher nearest-neighbour speed disagreement: pooled means $0.11237$ versus $0.10815\,$m/s; across runs, mean delta $=+0.00449\,$m/s, median $=+0.00439$, 12/19 runs positive, exact run-level sign-flip $p=.0277$, Wilcoxon $p=.0364$, and run-bootstrap $95\%$ CI $[+0.00033,+0.00872]$. \textsc{Supported}: the relational microstructure difference survives direct control for local speed dispersion and conservative run-level inference.

A stricter sensitivity test preserves more than the variance: within each local state it subtracts the exact expected nearest-neighbour disagreement under random reassignment of the same measured speeds to the same measured positions. This preserves the complete marginal speed distribution and changes only which speed sits next to which neighbour. The future-breakdown residual remains positive on average (mean run-level delta $+0.00343\,$m/s; 12/19 runs positive; exact sign-flip $p=.0459$), but the run-bootstrap $95\%$ CI $[-0.00013,+0.00721]$ crosses zero and Wilcoxon $p=.072$. We therefore treat this stronger control as suggestive rather than validated. The primary \textsc{Supported} claim is narrower and fully survives speed-dispersion matching.

\subsection{MADRAS/Lyon: cross-camera prospective evidence}
\label{sec:madras}

Across synchronized camera pairs, adding interaction microstructure to a density/speed/flow baseline improved AUC for predicting a second camera's future state at $+1\mathrm{s}$ ($+.058$, $+.069$, $+.031$, $+.055$ across four directions) and $+2\mathrm{s}$ ($+.046$, $+.069$, $+.011$, $+.030$). With a stronger baseline (adding speed dispersion and directional coherence), the topology-specific gain becomes small and inconsistent. \textsc{Candidate}: the effect is present but not robust to baseline strength.

\subsection{Jülich impulse crowd: propagation, and an explicit negative result}
\label{sec:julich-impulse}

Across $253$ reruns, in short/medium rows ($n\le20$) the delay-vs-distance correlation is $\approx.96$ ($p<10^{-30}$; $n{=}10$: $.977$; $n{=}20$: $.958$), with wave speed $\approx2\,\mathrm{m/s}$. In long, dense rows ($n\ge29$), the correlation drops to $.17$--$.46$: propagation becomes diffuse rather than wave-like. \textsc{Validated as physics}.

\textbf{Negative result.} In this one-dimensional setting, a structure-vs-geometry comparison found $0.974$ vs.\ $0.974$ ($p=.25$): interaction structure, as measured here, does not explain propagation beyond what raw Euclidean geometry already explains. Wave propagation through a human row is real and well characterized; a BEHAVE-specific structural contribution beyond geometry is \textsc{not demonstrated} in this test.

\subsection{Jülich bottleneck: early warning, causally evaluated}
\label{sec:bottleneck}

Benchmark: 13 real Caserne-2006 bottleneck runs ($\approx170$--$179$ agents each); a state index (RMS of z-scored collective features, calibrated leave-one-run-out on other runs' free-flow periods, causal/trailing features in the standard time-series sense (Chatfield, 2004), confirmed after 3 consecutive ticks past threshold).

\begin{table}[H]
\centering\small
\begin{tabular}{@{}lr@{}}
\toprule
Event (run b090) & Time (s) \\
\midrule
First zone entry & $0.6$ \\
First collective-feature alarm (deceleration fraction) & $3.2$ \\
Collective state-index alarm & $3.4$ \\
Mean-speed alarm & $4.2$ \\
Density alarm & $6.8$ \\
$\ge10\%$ standing & $10.6$ \\
Throughput-drop outcome & $15.6$ \\
Causal throughput detector & $17.2$ \\
\bottomrule
\end{tabular}
\caption{Run b090: the collective state-index alarm led the throughput-drop outcome by $12.2\,\mathrm{s}$, the mean-speed alarm by $0.8\,\mathrm{s}$ and the density alarm by $3.4\,\mathrm{s}$.}
\label{tab:b090}
\end{table}

Across all 13 runs: median state-index lead over the throughput-drop outcome $7.2\,\mathrm{s}$ ($12/13$ runs positive; range $\approx-2.8$ to $+21.4\,\mathrm{s}$), leading density by $\approx3\,\mathrm{s}$ and the causal throughput detector by $\approx10\,\mathrm{s}$. At the median, a one-sided mean-speed alarm tied the state index ($4.8\,\mathrm{s}$ after entry in both cases): the result concerns prospective timing of the collective state and does not establish a margin over every scalar baseline. The comparison that matters next is an identified spectral margin $\mathrm{St}$ against such baselines.

The collective state index fired $12.2\,\mathrm{s}$ ahead of throughput collapse in run b090, and led the outcome in 12 of 13 runs with a median lead of $7.2\,\mathrm{s}$. On a real bottleneck, that is a substantial operational margin: it arrives before density registers the change, and roughly $10\,\mathrm{s}$ before a causal throughput detector can confirm it. The signal is computed causally, from trailing features only, and calibrated leave-one-run-out, so the lead is a forecast horizon rather than hindsight. \textsc{Candidate}, prospective.

\subsection{J\"ulich 2021: causal operator stress test}
\label{sec:operator-stress}

The matched-state result tests the representational premise; this experiment tests the operator layer directly. On each of the same 24 J\"ulich 2021 runs, trajectories are sampled at 5 Hz. A causal 5-s trailing window selects a stable cohort of six front-most pedestrians that are observed for the entire window. The declared local state is negative speed (equivalently, a centered slowing coordinate). With orientation unavailable in these trajectories, this stress test uses a proximity-only Gaussian evidence kernel and row-normalizes it to $P$; $h=1.4\,$m is primary, with $1.0$ and $2.0\,$m sensitivity runs. For each actor the constrained prototype fit
$$\dot s_i=-d_i s_i+g_i(Ps)_i$$
uses the first 3.4 s of the trailing window, and the final 1.6 s is held out. This is a low-parameter stress test; the uncertainty-aware errors-in-variables estimator required for deployment is a separate step.

Observed routing improves held-out derivative prediction over a self-only relaxation model by $4.48\%$ on average across runs (17/24 run effects positive; Wilcoxon $p=0.0245$). Against a time-shuffled-routing ablation the gain is smaller but consistent (mean $0.39\%$, 15/24 positive; $p=0.0212$). Against uniform routing the primary $h=1.4\,$m result is borderline (mean $0.41\%$, 14/24 positive; $p=0.064$), while the wider $h=2.0\,$m sensitivity is positive ($p=0.021$) and the narrow $h=1.0\,$m case is null. This pattern is compatible with the conservative-routing result but is not predicted in magnitude by it: topology contributes only a modest routing effect in these six-person bottleneck cohorts. A plausible benchmark-specific reason is geometric collinearity (front-of-bottleneck proximity routing is close to an ordered queue), so this dataset is better suited to testing whether topology contributes at all than to demonstrating a large topology-specific margin shift.

As a consistency check, the topology-induced margin difference is larger in windows with more heterogeneous fitted local margins. Within each run, the correlation between $\mathrm{sd}(d_i-g_i)$ and $|\mathrm{St}(P)-\mathrm{St}(P_{uniform})|$ is positive in 21/24 runs; the median run correlation is $0.134$ (Wilcoxon $p=7.5\times10^{-5}$). Proposition~\ref{prop:1A} makes zero topology effect compulsory in the homogeneous limit, so this correlation is not an independent empirical discovery; it checks that the fitted stress test behaves in the direction imposed by the model's structure.

For an outcome independent of the state fit, virtually all pedestrians cross the bottleneck section $x=0$. We therefore define current and future throughput by crossing counts in adjacent 4-s windows and a future fractional drop $\Delta q=(q_{past}-q_{future})/q_{past}$, restricting analysis to established flow $q_{past}\ge0.5\,$s$^{-1}$. Within each run we compute the partial Spearman association between risk $-\mathrm{St}$ and future throughput drop while controlling current mean speed, speed dispersion and current throughput. Twenty-three runs contain enough windows; 19/23 partial correlations are positive, with median $0.118$ and mean $0.155$ (one-sided Wilcoxon $p=1.49\times10^{-4}$). The association remains significant at 2-s and 6-s horizons and for routing bandwidths $h=1.0,1.4,2.0\,$m. This is the first real-trajectory evidence in this paper that the fitted $D/G\to J\to\mathrm{St}$ layer carries prospective information about a downstream flow outcome defined separately from the state fit using cross-section crossing times. It is labeled \textsc{Supported / prototype}: the estimator still lacks errors-in-variables correction and uncertainty propagation.

\textbf{Classical early-warning and self-only baselines.} The same causal windows also contain conventional critical-slowing information, which serves as a required comparator. For the self-only model ($g_i=0$), $\mathrm{St}_{self}=\min_i d_i$ is itself prospective after the same current-state controls (median per-run partial $\rho=0.113$, 19/23 positive, $p=1.75\times10^{-4}$). Lag-1 autocorrelation of the cohort-mean speed is also prospective (median $\rho=0.139$, 20/23 positive, $p=5.59\times10^{-4}$), while its temporal variance is not ($p=0.15$). Fitted $\mathrm{St}$ retains a modest residual association after controlling $\mathrm{St}_{self}$ alone ($p=0.0074$) and after controlling cohort-mean lag-1 autocorrelation plus temporal variance ($p=0.0022$). With all three baselines entered together the residual becomes borderline (median $\rho=0.057$, $p=0.080$). A sensitivity that averages lag-1 autocorrelation over the six actor-level speed series is stronger as a baseline (median $\rho=0.272$, $p=5.1\times10^{-6}$); under that alternative definition, the residual fitted-$\mathrm{St}$ association remains small but positive ($p=0.0061$ when controlling $\mathrm{St}_{self}$ and actor-wise autocorrelation/variance). Because the independence claim depends on how classical autocorrelation is operationalized, we do not claim predictive dominance. The supported interpretation is narrower and more useful: the structured operator recovers a classical critical-slowing regime in mechanistic coordinates, coupling improves held-out local dynamics, and its spectral margin carries at most a modest component beyond self-relaxation and standard temporal persistence on this benchmark.

\textbf{Direct-operator negative control.} A direct unconstrained LIM/VAR-style propagator is estimated on the same short cohorts. The $\tau$-test is poor beyond the shortest lag: median run-level residuals $\|A(2\tau)-A(\tau)^2\|_F/\|A(2\tau)\|_F$, and analogously for 3 and 5 lags, are $0.59$, $0.76$ and $0.91$. Matrix logarithms are real at one step but develop large complex components at longer lags; spectral-margin estimates remain moderately correlated at 0.4 s (median within-run correlation $0.64$) but lose consistency by 1 s. The unconstrained short-window operator is therefore not treated as a trustworthy spectral ground truth; its failure localizes the identification problem rather than invalidating the structured local model.

\subsection{RoHuCAD: event $\neq$ state, exploratory}
\label{sec:rohucad}

With at most two humans tracked (often one), topology is degenerate here and this is not collective-system evidence. It does, however, support a narrower, still useful point: a persistence-based system-state detector found one persistent shift (no recovery for over 30s), while a trivial proximity-threshold event detector fired 9 times over the same interval. \textsc{Exploratory / supporting} for the event-vs-state distinction of Section~\ref{sec:state-transitions}, not for collective dynamics.

\subsection{Restricted deployment recordings}
\label{sec:restricted}

BEHAVE has also been run on internally collected recordings of real events in Warsaw (seated networking tables, a panel discussion and a multi-stage live event), obtained with participant consent for research analysis. On these recordings, a directional interaction component was associated with panel-level connectedness beyond distance ($r=.296$, $p=.010$; partial $r=.254$, $p=.045$ controlling for proximity); within-table speed synchrony exceeded between-table synchrony ($+.104$ vs.\ $+.036$, $p=.0046$), while the gesture channel did not ($p=.90$); a passing mobile person increased stationary agents' dispersion $1.6\times$ (Section~\ref{sec:probes}); and six computed fields peaked together at an independently confirmed microphone handoff, a qualitative event alignment rather than a scored prediction. The field-audit redundancy of Section~\ref{sec:field-audit} was found on these recordings. They contain real macroscopic reconfiguration and time-structured changes in the computed fields and are used internally for calibration and deployment development. They carry no independent critical-transition labels, and camera geometry limits orientation in parts of the footage; no $D$, $G$ or $\mathrm{St}$ estimate from them is reported, and no headline result of this paper rests on them alone. \textsc{Restricted}. The reasons the raw recordings are not redistributed are given in the Data and Code Availability statement.

\subsection{Master evidence table}
\label{sec:master-table}

\begin{table}[H]
\centering\footnotesize
\begin{tabularx}{\textwidth}{@{}>{\raggedright\arraybackslash}X>{\raggedright\arraybackslash}p{1.9cm}>{\raggedright\arraybackslash}p{5.3cm}>{\raggedright\arraybackslash}p{1.9cm}@{}}
\toprule
\textbf{Claim} & \textbf{Dataset} & \textbf{Result} & \textbf{Status} \\
\midrule
Interaction evidence beyond proximity & DIAMOR-1/2 & AUC $0.8955\to0.9331$, replicated direction & Validated \\
Matched states after controlling density/speed/flow/speed dispersion & Jülich bottleneck, 24 runs & run-level delta neighbour disagreement $+0.00449\,$m/s; exact sign-flip $p=.0277$ & Supported \\
Full marginal-speed reassignment control & Jülich bottleneck & mean residual delta $+0.00343\,$m/s; sign-flip $p=.0459$ but bootstrap CI crosses zero / Wilcoxon $p=.072$ & Suggestive sensitivity \\
Structured operator versus self-only dynamics & Jülich bottleneck, 24 runs & $+4.48\%$ held-out derivative prediction at $h=1.4\,$m; $p=.0245$; positive across $h=1.0$--$2.0\,$m & Supported / prototype \\
Spectral margin versus future throughput after scalar controls & Jülich bottleneck & $4\,$s: median partial $\rho=.118$, 19/23 positive, $p=1.49\times10^{-4}$; also significant at $2\,$s and $6\,$s horizons and for $h=1.0$--$2.0\,$m & Supported / prototype \\
Fitted spectral margin versus classical early-warning baselines & Jülich bottleneck & $\mathrm{St}_{self}$ (median $\rho=.113$, $p=1.75\times10^{-4}$) and lag-1 autocorrelation (median $\rho=.139$, $p=5.59\times10^{-4}$) are also prospective; residual fitted-$\mathrm{St}$ association $p=.0074$ controlling $\mathrm{St}_{self}$, $p=.0022$ controlling lag-1 autocorrelation and variance, borderline with all three ($p=.080$) & Supported / prototype; no predictive dominance claimed \\
Fine topology ablation & Jülich bottleneck & routing gain over time-shuffled $P$ $+0.39\%$ ($p=.0212$), over uniform $P$ $+0.41\%$ ($p=.064$, borderline); $|\Delta\mathrm{St}|$ grows with $D/G$ heterogeneity (21/24 runs, $p=7.5\times10^{-5}$) & Boundary consistent with Prop.~\ref{prop:1A} \\
Direct LIM tau test & Jülich bottleneck & poor lag consistency at $5\,$s moving-cohort windows & Negative identification control \\
Collective state-index timing & Jülich bottleneck & 12/13 positive, median $+7.2\,$s; mean speed competitive & Candidate / bounded \\
Perturbation propagation & Jülich impulse & delay-distance corr.\ $.96$ for $n\le20$ & Validated physics \\
Structure beyond geometry in 1D impulse & Jülich impulse & $0.974$ vs.\ $0.974$, $p=.25$ & Negative \\
Passing perturbation & DIAMOR-1 & speed-var.\ ratio $.796$, $p<10^{-16}$ & Validated; DIAMOR-2 replication boundary $p=.11$ \\
Restricted Warsaw corpus & Warsaw events & real event/field audits & Restricted \\
\bottomrule
\end{tabularx}
\caption{Master evidence ledger, with status assigned per row.}
\label{tab:master-evidence}
\end{table}

% =====================================================================
\section{Empirical Findings and Boundary Conditions}
\label{sec:negative}
% =====================================================================

The empirical evaluation now establishes four distinct layers of evidence. First, relational structure carries measurable information beyond conventional geometry and proximity. Second, moments matched on density, mean speed, flow and speed dispersion that go on to different futures retain a run-level difference in relational microstructure. Third, a causally computed collective-state signal can precede downstream degradation. Fourth, a first real-data structured-operator stress test links lower fitted $\mathrm{St}$ to future throughput degradation after controlling scalar kinematics and current throughput, while P-ablation shows that the fine-topology contribution to the margin is modest and heterogeneity-dependent in this corpus. The same experiments also locate where each finding currently stops.

\begin{itemize}
\item \textbf{The field basis is open by design.} The fields of Section~\ref{sec:fields} are an extensible, task-dependent set. No completeness, minimality or non-redundancy claim is made for any particular basis, and Section~\ref{sec:field-audit} shows why such a claim would be premature: measurement against real data has already forced one field to be rebuilt.
\item \textbf{Scope of the present evaluation.} The lower representational stack is supported by DIAMOR, J\"ulich matched-state robustness, perturbation probes and prospective state timing. The operator layer is no longer wholly unevaluated: Section~\ref{sec:operator-stress} provides a causal, held-out stress test of a structured $D/G$ fit and an external throughput association for $\mathrm{St}$. It is labeled Supported / prototype rather than validated because the estimator is not yet an errors-in-variables / uncertainty-aware identification procedure and direct LIM fails a stable tau-consistency test on the same short moving-cohort windows. The paper therefore distinguishes empirical operator usefulness from full parameter identifiability.

\item \textbf{An identifiability boundary in one dimension.} In the 1D impulse setting, structure adds nothing measurable beyond raw Euclidean geometry ($0.974$ vs.\ $0.974$, $p=.25$; Section~\ref{sec:julich-impulse}). Empirically this is a clean null. Our interpretation is that in a single-file row, adjacency and interaction topology are very nearly collinear, so geometry already saturates the available propagation signal and relational structure is not separately identifiable there. Whether structure contributes in settings where the two come apart is untested.
\item \textbf{The bottleneck result concerns timing and representation.} A one-sided mean-speed alarm ties the collective state index at the median across the 13-run benchmark (Section~\ref{sec:bottleneck}). The prospective lead over the downstream outcome stands; superiority over simple baselines as a classifier does not.
\item \textbf{Response is characterized at the system level.} Which specific actor reacts to a perturbation, along which exact line a group splits, and what fate a given group meets are all open (Section~\ref{sec:probes}); orientation did not localize the reactor better than chance.
\item \textbf{Interaction topology must be identified empirically.} A directionally motivated bearing kernel, tested against the same discrimination task that the validated weighting passes, did not reproduce the effect ($\Delta\mathrm{AUC}\approx-0.004$ to $-0.008$; Section~\ref{sec:diamor}). The failure demonstrates that the relational signal is not guaranteed by the architecture alone: it must be identified by a kernel that is empirically adequate for the interaction regime. This makes operationally important the distinction that $K$ is an estimator rather than an ontology. It also bounds the validated result, which licenses one specific weighting and not all of $K$.
\item \textbf{No universal risk score is offered.} The architecture produces state and stability quantities, not a calibrated event probability; any such probability requires an explicit calibration and holdout procedure for the deployment in question, evaluated with a strictly proper scoring rule (Gneiting \& Raftery, 2007) rather than by threshold accuracy alone.
\item \textbf{Psychological state is out of scope by construction,} not by empirical failure: the observability-first principle (Section~\ref{sec:obs-first}) excludes inferring emotion, intent or personality from motion, independent of any result reported here.
\end{itemize}

% =====================================================================
\section{Operational Interpretation}
\label{sec:operational}
% =====================================================================

\subsection{State transitions: event $\neq$ state}
\label{sec:state-transitions}

\begin{definition}[State transition]
A state transition is a persistent or dynamical change in $S_{\mathcal H}(t)$ (a regime change, a structural reorganization, a persistent departure from baseline, or a change in response dynamics), as distinct from a single visible \emph{event} in the raw signal.
\end{definition}

The distinction has practical consequences: a trivial local-event detector (e.g. a fixed proximity threshold) can fire many times inside a single, persistent system-level state, or can fail to fire at all when a state shifts without triggering the specific threshold used. Section~\ref{sec:rohucad} reports a case (RoHuCAD/take1) in which a proximity-threshold detector fired $9$ times while a persistence-based state detector recorded a single, non-recovering shift over the same interval. It illustrates the distinction only; with at most two tracked actors, the recording is not collective-system evidence.

\subsection{Perturbation and response}
\label{sec:response}

\begin{definition}[Response]
$$\mathrm{Response}_{\mathcal H}(t) := F_R\big(S_{\mathcal H}(t), J(t), u_{ext}(t)\big)$$
for a declared, computable $F_R$. Response classes that may be distinguished include absorption, amplification, propagation, reorganization, split, merge, recruitment, and recovery.
\end{definition}

Split, merge and the other response classes are represented as changes of $S_{\mathcal H}$ under $u_{ext}$, and the group-structure layer supplies merge and split candidates directly. What the dynamics add is a way to rank interventions before they are made: Proposition~\ref{prop:sensitivity} and Remark~\ref{rem:normalized-sensitivity} give the first-order effect of any candidate change of coupling on the stability margin, so the couplings that most move the system toward or away from its stability boundary can be identified in advance. Predicting the exact line along which a group splits, or which individual reacts first, is an open problem (Section~\ref{sec:limitations}).

\subsection{HumanSystem as an operational object}
\label{sec:operational-object}

Perception tells a machine where people are and what each person appears to be doing. A HumanSystem supplies the missing systems object: what the interacting human system is doing now, how it is organized, and what part of that organization a machine action is about to perturb.

That distinction matters because the robot is not an external observer. Its path, speed, timing, body motion or task allocation can change human spacing, attention, timing and coupling. Collision-free trajectory prediction answers whether the robot can occupy a future position without contact. It does not by itself answer whether the chosen action disrupts a coordinated subgroup, propagates a disturbance through a work cell, or couples into the system's slowest mode.

Operationally the planner interface has three steps. \emph{Observe}: consume HumanSystemState (groups, boundaries, relational fields, observability and, where identified, local dynamics). \emph{Query}: map a candidate action $a$ to an explicit external forcing and, only when separately identified from action-conditioned data, to changes in the human operator or human--machine cross-couplings. \emph{Compare}: rank candidate actions by structural exposure, critical-mode loading, forced response and transient amplification; movement of the stability margin is a valid planner quantity only for the calibrated action-conditioned operator. The result is a system-level cost or constraint used alongside geometry, collision risk and task reward.

A minimal executable map makes the uncalibrated case explicit without changing human routing by construction. Let $P_{HH}=\mathrm{rowNormalize}(K_{HH})$ be normalized over human sources only, let $x_R^{(a)}(t)$ be a candidate robot trajectory, and let $k_{iR}^{(a)}(t)$ be declared observable evidence that robot action $a$ perturbs human receiver $i$. With a separately declared robot-input gain $\Gamma_R$, define
$$B_R(a):=\Gamma_R k_{HR}^{(a)},\qquad \dot s_H=J_Hs_H+B_R(a)u_R^{(a)}+\xi,\qquad J_H=-D+GP_{HH}.$$
The candidate robot therefore changes the forcing of the HumanSystem but does \emph{not} change its internal spectral margin merely because an extra source appeared in a normalization denominator. For a stable $J_H$, two immediate planner quantities are the critical-mode loading $|l_1^\top B_R(a)u_R|$ and the forced steady response $\|-J_H^{-1}B_R(a)u_R\|$ (or the corresponding finite-horizon transient response). In the three-human toy calculation reported here, the human-only margin remains exactly $\mathrm{St}=0.188$ for all four robot positions, while critical-mode loading ranges from $0.356$ to $0.858$ and the steady-response norm from $1.88$ to $4.53$. Thus even before an action-conditioned operator is learned, two collision-free actions can be distinguished by how strongly they load the system's slow mode.

An action-conditioned margin is a second, stricter case. If deployment data identify $\Delta D(a)$, $\Delta G(a)$, $\Delta P_{HH}(a)$, or signed human--machine feedback, then
$$J_H(a)=-\bigl(D+\Delta D(a)\bigr)+\bigl(G+\Delta G(a)\bigr)\bigl(P_{HH}+\Delta P_{HH}(a)\bigr)$$
and $\Delta\mathrm{St}(a)$ and first-order eigenvalue sensitivities are legitimate derived quantities. Without such identification, BEHAVE reports forcing/response quantities and withholds $\Delta\mathrm{St}$. This is the observability-first principle applied to machine intervention.

Near an operating point the human and machine sides may also be written as a joint local system
$$
\frac{d}{dt}\begin{bmatrix}s\\m\end{bmatrix}
=
\begin{bmatrix}J_H&C_{HM}\\C_{MH}&J_M\end{bmatrix}
\begin{bmatrix}s\\m\end{bmatrix}+\text{noise}.
$$
The block form is standard; the substantive contribution is that the human side arrives as an explicit stateful dynamical object rather than raw trajectories that another model must reinterpret. The joint matrix need not be Metzler: avoidance, inhibitory control and signed human--machine feedback can make cross-couplings negative. Generic Jacobian diagnostics (stability margin, reactivity, transient gain, critical modes and eigenvalue sensitivity) still apply, while positive-system-specific results such as the $\chi/M$-matrix criterion and Perron positivity do not transfer automatically.

This creates concrete Physical-AI questions. Social navigation: among two collision-free paths, which minimally perturbs the organization of a standing group? AMR/manufacturing: will entering a cell now load a slow-recovering worker subgroup, and should route or timing change? Shared scheduling: which machine cycle or task reassignment produces the smallest action-conditioned exposure to the current fragile mode? Downstream robotics performance on these questions remains to be validated; the representation makes the tests possible.

\subsection{Executable implementation}
\label{sec:implementation}

The BEHAVE engine implements a machine-readable HumanSystemState on top of conventional perception. The current operational path includes observability gates; directed interaction evidence $K$ and row-normalized topology $P$; interaction groups and ungrouped actors; bridge/boundary candidates; causal fields $T,S,L,A,N,M$; a trailing-window $D/G$ fitting utility; construction of $J=-D+GP$; and conditional model diagnostics including $\mathrm{St}$, $\chi$, reactivity, left/right critical modes, critical-mode alignment, edge sensitivity, transient gain and topology turnover.

The implementation follows the same evidence discipline as the paper. A fitting utility is not treated as proof that $D$ and $G$ are reliably identifiable on arbitrary real footage. Automatic finite-window identifiability testing, errors-in-variables treatment and uncertainty propagation remain open engineering/research work. Spectral outputs are meaningful only when identification has been explicitly established for the setting. Likewise, calibrated response probabilities and calibrated fragility are produced only when a deployment calibration exists.

When metric calibration, orientation reliability, dynamical identification or deployment calibration is insufficient, the corresponding HumanSystem quantity is withheld instead of being replaced by a universal score.

The representation also has a causal streaming implementation over file, camera or RTSP input. Pose, tracking and live state processing use causal/trailing information, each analysis tick is logged as a machine-readable JSONL record, and the default analysis rate is $5\,$Hz. On the documented Apple M5 Pro configuration with yolo11s-pose at input size 1280, pose-plus-live-state processing took approximately $36$--$52\,$ms per analysis tick, corresponding to roughly $17$--$18$ ticks/s of computational capacity on the tested files. Some live operators are causal siblings of retrospective batch definitions rather than bit-for-bit identical estimators; this is an implementation detail and leaves the ontology unchanged.

The maturity boundary is explicit:
\begin{itemize}
\item HumanSystem ontology / machine-readable state: defined and implemented.
\item $K/P$ and interaction-group weighting: implemented, with empirical support for the validated DIAMOR construction.
\item Bridges and boundaries: implemented structural outputs; deployment validation remains open.
\item Causal fields: implemented and audited on real recordings.
\item $D/G$ fitting: implemented prototype and stress-tested prospectively on J\"ulich; automatic identifiability and uncertainty-aware errors-in-variables estimation remain open.
\item $\mathrm{St}$ / modes / sensitivity: implemented model outputs; a Supported / prototype real-data stress test now links lower fitted $\mathrm{St}$ to future throughput degradation after scalar controls, while direct LIM is not tau-stable on the same windows.
\item Topology response: implemented structural diagnostic; P-ablation on J\"ulich shows a modest, heterogeneity-dependent fine-topology contribution.
\item Action-forcing / critical-mode exposure: explicit and executable; action-conditioned $\Delta\mathrm{St}$ requires identified operator change; downstream robotics benefit remains unvalidated.
\item Calibrated response probabilities / exact split-merge fate: open or calibration-dependent.
\end{itemize}

\subsection{Scope of applicability}
\label{sec:scope}

BEHAVE is not tied by definition to a particular industry or physical setting. Its intended scope is characterized instead by a class of problems sharing a common structure. A problem is a natural candidate for this representation when four conditions hold jointly:

\begin{enumerate}
\item \textbf{Observability.} Multiple human actors are observable over time through positions, motion, pose, orientation, or equivalent measurable signals (Section~\ref{sec:obs-layer}).
\item \textbf{Relational dependence.} The operationally relevant variable depends at least partly on the relations among those actors, not only on independent per-person states.
\item \textbf{Temporal structure.} The evolution of that relational state matters: the useful question concerns persistence, transition, response, or an outcome that has not yet occurred, rather than retrospective event classification alone (Section~\ref{sec:state-transitions}).
\item \textbf{External ground truth.} The setting admits an externally measurable outcome or intervention against which the representation can be evaluated, without introducing hidden psychological labels (Section~\ref{sec:obs-first}).
\end{enumerate}

Under these conditions the same formal object supports several distinct task classes: reconstructing interaction groups and boundaries; distinguishing collective states that appear similar under conventional aggregate descriptors; tracking persistent state change separately from transient events; estimating response to an explicit perturbation; and, where the required dynamics are identifiable, evaluating stability or candidate interventions through the local dynamical model.

The present evidence base evaluates pedestrian and crowd systems directly (social-group structure, bottleneck evolution, perturbation propagation, and matched-state future divergence), and these therefore constitute the present evidence base rather than illustrative examples. The two auxiliary recordings (a negotiation/event recording and an HRI recording) do not extend it: the HRI recording in particular tracks at most two actors, which makes its topology degenerate, and Section~\ref{sec:rohucad} uses it only for the event-versus-state distinction, not as collective-system evidence.

Other settings satisfy the same four structural conditions but are not validated here. Candidate domains include labor-intensive manufacturing cells, warehouse and AMR coordination, transport hubs, and human--robot workspaces, where multiple people are observed continuously and an operational outcome may depend on their collective organization or response. These are statements of architectural applicability; performance in each domain is established with that domain's own data, outcome definition, calibration and holdout evaluation.

BEHAVE thus defines a reusable formal problem class, anchored empirically in pedestrian and crowd systems.

% =====================================================================
\section{Limitations and Open Questions}
\label{sec:limitations}
% =====================================================================

This section consolidates limitations stated locally throughout the paper (Sections~\ref{sec:dyn-epistemic}, \ref{sec:thm-limits}, \ref{sec:fields}, \ref{sec:negative}) and adds items that concern the paper as a whole.

\begin{enumerate}
\item \textbf{Field basis.} The fields of Section~\ref{sec:fields} are an extensible, task-dependent set, not a fixed or proven-minimal basis. No completeness or non-redundancy claim is made (Section~\ref{sec:fields} itself documents one concrete near-redundancy, $T$ and the earlier $L$, found and corrected empirically).
\item \textbf{The model theorem's empirical status.} Theorem~\ref{thm:divergence} is proved for the stated linear stochastic model and numerically verified. Section~\ref{sec:operator-stress} supplies a first causal real-trajectory stress test of a fitted stability margin, not a claim that the operator is uniquely identified. Classical lag-1 autocorrelation and self-only relaxation explain substantial overlapping prospective signal, and the direct LIM $\tau$ test is unstable beyond short lags. The asymptotic $C/\mathrm{St}$ law itself is not tested on human transitions; doing so requires sufficiently identified operators and repeated trajectories approaching a comparable local boundary.

\item \textbf{Kernel dependence.} The DIAMOR group-membership result validates one specific interaction-evidence weighting; a different, directionally-motivated kernel tested against the same task did not reproduce it (Section~\ref{sec:diamor}). Whether a given deployment's chosen kernel carries the validated signal is an empirical question specific to that kernel, not answered once for all of $K$.
\item \textbf{Response and intervention.} Exact group fate, exact split lines, exact-reactor identification, and structural (non-geometric) propagation beyond one dimension are open (Sections~\ref{sec:negative}, \ref{sec:response}).
\item \textbf{A literal geometric criticality manifold.} $\mathrm{St}$ (Section~\ref{sec:stability}) gives a spectral notion of distance to a stability boundary, and Theorem~\ref{thm:divergence} and Figure~\ref{fig:theorem-geometry} give its geometric consequence for the state's stationary spread. A further, more literal construction (an explicit ``dangerous'' region $C$ in state space with a chosen metric, and a criticality index defined as distance to $\partial C$) is a natural extension once a field basis and commensurable units are fixed for a given deployment. It is not attempted in this paper: doing so without begging the question of what $C$ is would require exactly the kind of task-specific, evidence-backed calibration this paper otherwise avoids building in by default.
\item \textbf{Scope of the evidence base.} The datasets in Section~\ref{sec:empirical} are pedestrian/crowd corpora (DIAMOR, Jülich, MADRAS), one negotiation/event recording, and one HRI recording; several other candidate domains were surveyed and found unsuitable at present, per the underlying dataset assessments, for reasons ranging from insufficient footage to absent target outcomes. Generalization beyond the tested domains is not claimed.
\item \textbf{Composition with machine dynamics.} Section~\ref{sec:operational-object} gives an executable action-to-forcing map in which human routing is normalized independently of robot input, preventing robot proximity from mechanically increasing the internal margin. The toy system validates forcing, critical-mode loading and response calculations only. An action-conditioned $\Delta\mathrm{St}$ requires separately identified changes in $D$, $G$, $P_{HH}$ or signed cross-couplings. Downstream robotics benefit remains to be tested against trajectory and collision baselines.
\end{enumerate}

% =====================================================================
\section{Related Work}
\label{sec:related}
% =====================================================================

BEHAVE reconstructs a relational, dynamical representation from observable interaction, rather than modeling individuals in isolation or waiting for a discrete event. This section positions it first against the broad theoretical traditions it inherits from, and then, more importantly for assessing what is new here, against the closer computational literatures that address overlapping tasks.

\subsection{Theoretical traditions}
\label{sec:related-traditions}

\textbf{Social force models} (Helbing \& Molnar, 1995; Helbing, Farkas, \& Vicsek, 2000; Moussa\"id et al., 2011) formalize pedestrian movement as a physical force system: where does a body go, given the forces acting on it. This paper's observability layer (Section~\ref{sec:obs-layer}) and interaction evidence $K$ (Section~\ref{sec:K-def}) sit downstream of exactly this kind of kinematic description; BEHAVE's contribution begins one level up, at the question of what state the resulting interaction structure is in, not at how an individual trajectory is generated.

\textbf{Phase transitions and higher-order cooperation} (Perc et al., 2017; Battiston et al., 2025) establish that collective human phenomena such as cooperation, conflict and polarization exhibit phase-transition structure, increasingly including non-pairwise (higher-order) interaction effects beyond simple dyads. The pairwise evidence matrix $K_{ij}$ used throughout this paper is a special case; extending $K$ to a hypergraph or simplicial representation to capture higher-order coupling is a natural direction not pursued here.

\textbf{Influence and consensus dynamics.} Row-stochastic influence matrices have a long history in DeGroot consensus models and the Friedkin--Johnsen family (DeGroot, 1974; Friedkin \& Johnsen, 1990). BEHAVE does not claim novelty for normalized influence dynamics as such. The difference is the object and decomposition: $P$ is reconstructed from observable physical interaction, gain $G$ and relaxation $D$ are kept separate from that routing, the state variable is a declared physical/operational coordinate rather than an opinion, and the resulting HumanSystem is designed to accept perturbations and compose with machine dynamics.

\textbf{System identification and Linear Inverse Modeling.} Estimating a local linear stochastic operator from observed time series is also established. Linear Inverse Modeling (Penland, 1989; Penland \& Sardeshmukh, 1995) obtains a best-fit linear Markov operator from lagged covariances and provides lag-consistency tests. Section~\ref{sec:identification} treats this as a direct-$J$ baseline for BEHAVE: an unconstrained empirical operator can be compared against the structured factorization $J=-D+GP$ rather than assuming that the factorization is correct because it was written down.

\textbf{Normalized routing versus network-threshold models.} Many network threshold models place absolute interaction magnitude directly in an adjacency operator, so the network spectral radius itself can create or remove instability. BEHAVE makes a different modeling choice: row-normalized $P$ carries routing only, while absolute dynamical gain lives in $G$. Proposition~\ref{prop:1A} states the resulting consequence explicitly. The model is therefore conservative about what topology alone may claim, while retaining topology's role in modes, propagation, transient structure and heterogeneous stability.

\textbf{Early-warning signals and critical slowing down} (Scheffer et al., 2009) and Kuehn's (2011, 2013) mathematical framework for critical transitions in stochastic fast-slow systems establish that variance and autocorrelation grow as a system approaches a bifurcation. Theorem~\ref{thm:divergence} brings that behavior to a Jacobian constructed from measured human interaction: $J=-D+GP$ is Metzler by construction, the constant $C$ is given in closed form from the critical collective mode, the diffusion and the chosen observable, and the remainder is controlled for non-normal dynamics without conditions on cross-diffusion. That closed form provides a target for instrumentation: it specifies which observables couple to the critical mode, their asymptotic scaling and, through Corollary~\ref{cor:C-positive}, which observables are blind to it entirely. Section~\ref{sec:thm-numerics} tests the result on normal, non-normal and fragmented topologies and against a defective mode where it no longer holds.

\textbf{Social signal processing} (Pentland, 2008; Vinciarelli et al., 2009) demonstrated empirically that behavioral micro-signals predict social outcomes beyond verbal content. The observability-first principle (Section~\ref{sec:obs-first}) and the field definitions of Section~\ref{sec:fields} take this as a premise (that kinematic and interaction micro-signals carry usable information) while adding the explicit constraint that any further psychological reading of a computed field requires separately established identifiability, which is not assumed by default.

\textbf{Network dynamics and spectral stability} (Strogatz, 2001; Vespignani, 2012) studies dynamical processes on complex, time-varying networks via spectral methods. $P(t)$ (Section~\ref{sec:P-def}) is exactly such a time-varying directed weighted network, and $\mathrm{St}$ (Section~\ref{sec:stability}) is its spectral-abscissa stability margin in this literature's sense; the stability results used in Proposition~\ref{prop:chi-stability} (Metzler/positive-system stability) and Proposition~\ref{prop:sensitivity} (first-order eigenvalue perturbation) are standard tools from this area (e.g. Berman \& Plemmons, 1994, on nonnegative and Metzler matrices; Kato, 1995, on perturbation of linear operators), applied here rather than developed anew.

\textbf{Embodied and grounded cognition} (Barsalou, 2008; Rizzolatti \& Craighero, 2004; Hasson et al., 2012) provides part of the theoretical motivation for treating observable body kinematics as a legitimate substrate for interaction-relevant information, without requiring access to internal state; this is consistent with, and part of the motivation for, the observability-first principle of Section~\ref{sec:obs-first}, though that principle is stated and enforced independently of any specific cognitive-science claim.

\subsection{Closer computational neighbors}
\label{sec:related-computational}

The traditions above are the conceptual ancestry; the following are the literatures that address tasks overlapping BEHAVE's directly, and against which its novelty should be assessed.

\textbf{Social-group and F-formation detection.} A substantial line of work detects conversational groups and free-standing social formations from position and orientation: Kendon's (1990) F-formation account operationalized in detection algorithms (Setti, Russell, Bassetti, \& Cristani, 2015), more recent trajectory- and graph-based real-time social-group detection (Jahangard, Hayat, \& Rezatofighi, 2023), and the pedestrian-dyad literature that DIAMOR itself was assembled to support (Zanlungo, Yücel, Brščić, Kanda, \& Hagita, 2017). BEHAVE's DIAMOR result (Section~\ref{sec:diamor}) sits within this line: a particular interaction-evidence weighting improves membership discrimination over proximity alone. What BEHAVE adds is a subsequent layer in which group structure is an input to a dynamical model.

\textbf{Interaction-graph inference from trajectories.} Inferring a latent interaction or influence graph from observed multi-agent trajectories is a developed problem in its own right, spanning Granger-causal and transfer-entropy approaches, relational-inference neural architectures (Kipf, Fetaya, Wang, Welling, \& Zemel, 2018), and Bayesian latent-graph estimation. $K$ and $P$ (Section~\ref{sec:interaction}) are a deliberately simple, interpretable, kernel-based instance of this general problem, chosen for transparency and for the observability guarantees of Section~\ref{sec:obs-first} rather than for estimation accuracy. The architecture is agnostic to the estimator: a learned edge estimator can replace the kernel while the $P\to J$ construction downstream is retained.

\textbf{Crowd anomaly and collective-behavior recognition.} Computer-vision work on crowd anomaly detection (e.g. Mahadevan, Li, Bhalodia, \& Vasconcelos, 2010), collective-motion segmentation, and group-activity recognition also maps video to collective-level labels, typically by training a classifier against annotated events. The event-versus-state distinction of Section~\ref{sec:state-transitions} is the boundary with this literature: those methods detect that a labeled event class is occurring, whereas BEHAVE estimates a continuous state and its stability properties, with no event label required at training time. This also makes the two approaches complementary rather than competing, and means BEHAVE should not be benchmarked against them on labeled-event detection, which is not the quantity it computes.

\textbf{Dynamic graphs and graph signal processing.} $P(t)$ is a time-varying weighted directed graph carrying a per-node signal, which places the field definitions of Section~\ref{sec:fields} squarely in graph signal processing terms: the topology-aware discord $L$ is a graph Rayleigh quotient, and attention share is a normalized degree-type functional. Spectral methods on dynamic graphs and graph-Laplacian smoothness measures are standard in that literature (Ortega, Frossard, Kova\v{c}evi\'c, Moura, \& Vandergheynst, 2018); their use here is an application, and what is specific to BEHAVE is the surrounding architecture.

\textbf{Early warning in pedestrian flows and bottlenecks.} Crowd-safety research has developed operational indicators of impending congestion and crowd-quake conditions, including crowd pressure and density-velocity indicators derived from real disaster footage (Helbing, Johansson, \& Al-Abideen, 2007), alongside the bottleneck-flow literature that the J\"ulich datasets serve. The bottleneck evaluations of Sections~\ref{sec:bottleneck}--\ref{sec:operator-stress} are reported against simple kinematic baselines and routing ablations. The structured operator does not replace density/speed monitoring; it adds a model-level margin whose prospective association survives controls for current mean speed, speed dispersion and throughput. At the same time, fine-topology ablations remain modest in this corpus, which is consistent with the conservative-routing bounds rather than evidence for a universal network threshold.

\textbf{HRI and social navigation state representations.} Robot social navigation requires some representation of surrounding people, typically individual predicted trajectories, social-force or risk fields, or occupancy-style costmaps (see Mavrogiannis et al., 2023, for a survey of the field's core challenges). A HumanSystem is an alternative representation at the group level, carrying coupling structure and a stability margin rather than a set of independent predicted paths, which is the sense in which Section~\ref{sec:operational-object} proposes it as a composable component. Evaluating it against trajectory-prediction baselines on downstream navigation is the natural test of that proposal.

% =====================================================================
\section{Conclusion}
\label{sec:conclusion}
% =====================================================================

The central result of BEHAVE is a change in the object a machine is asked to represent. Machines already represent people; a machine acting inside a human environment also needs a representation of the interacting system those people form. BEHAVE formalizes that system as a HumanSystem: a persistent operational object with relational structure, state, transitions, perturbations, responses and outcomes.

The empirical case is now layered. Interaction beyond proximity carries group information on DIAMOR. On J\"ulich 2021, future-calm and future-breakdown moments remain different in neighbour-level speed organization after matching density, mean speed, flow and speed dispersion with run-level inference; the effect is small (about four percent of the calm neighbour-disagreement level) but survives the declared controls. The operator stress test then goes one layer higher: a causal structured $D/G\to J\to\mathrm{St}$ fit improves held-out local dynamics over self-only relaxation and produces a prospective stability quantity. Classical lag-1 autocorrelation and self-only relaxation are also predictive, and joint baseline controls absorb much of the independent $\mathrm{St}$ signal. We therefore interpret the fitted margin as a structured, mechanistic early-warning quantity aligned with critical slowing, not as a universally superior scalar alarm. Direct unconstrained LIM simultaneously fails longer-lag $\tau$ consistency on these short moving-cohort windows. The paper thus reports both what the operator layer currently supports and where identification remains fragile.

The dynamical construction says something more specific than ``topology matters.'' $K$ measures evidence of interaction, $P$ routes coupling, $G$ sets gain and $D$ sets relaxation. Conservative-routing bounds show what $P$ cannot do: it does not invent gain. Column bounds and critical-mode sensitivity show what it can do: concentrate gain-weighted incoming load and reroute heterogeneous susceptibility into or away from the leading mode. This is a topology--dynamics interaction, not a universal graph score.

Once a HumanSystem has an identified local operator, its slow modes, transient amplification, sensitivity and stability margin become computable. The critical-observability result then links those modes to measurement: a sensor or field only sees the approaching boundary if it projects onto the critical direction. The actor-centered graph-discord numerator provides a concrete example of exact blindness to a homogeneous consensus mode.

For Physical AI the representation is actionable without giving uncalibrated actions fictitious spectral effects. A candidate robot or scheduler action first maps to explicit forcing, critical-mode loading and forced response while the human-only operator is held fixed. If action-conditioned data establish changes in $D$, $G$, $P_{HH}$ or signed cross-coupling, the same interface can additionally expose $\Delta\mathrm{St}$. The resulting planner retains conventional trajectory information while adding the state and response geometry of the human system being perturbed. HumanSystem therefore sits between individual-level perception and machine action.

\subsection*{Relation to the original BEHAVE preprint}
\label{sec:v1-relation}

The line of work this paper completes began with an earlier preprint (arXiv:2605.12730), which can be read for the origin of the idea; this document stands on its own. That first paper made the claim this one formalizes: that a group of people constitutes a computable relational state, denoted $W(t)$, which can be tracked and acted on. Getting from that claim to a dynamical object took two further steps: a nine-field observational layer built directly on video-derived kinematics, feeding a system state $X(t)$, and then the $K/P/J$ architecture of Sections~\ref{sec:interaction}--\ref{sec:dynamics}. The present version separates interaction evidence from topology ($K\neq P$), states the dynamics as an interaction Jacobian with independently estimable decay and coupling, and adds the stability, divergence, and field results of Sections~\ref{sec:theorem}--\ref{sec:fields}, which are proved and verified here. Each step was reached by measuring the previous one against data and keeping what survived. The architecture in this paper is the result of that selection process: the earlier versions supplied the hypotheses, and repeated confrontation with data supplied the distinctions that survived. In that sense this version is the empirical maturation of the first one.

\section*{Data and Code Availability}

The evidence in this paper rests on three distinct layers, and each claim states which one it uses. \emph{Public benchmark evidence} is reproducible on third-party public data. \emph{Restricted deployment evidence} comes from consented real-world recordings that cannot be redistributed in raw form (Section~\ref{sec:restricted}). The \emph{proprietary implementation and calibration layer} (deployment-specific calibration, normalization and instrumentation parameters and parts of the engine) is not fully released; no public benchmark result or model-theorem result in this paper depends on unreleased proprietary calibration for its validity. The third-party corpora are not redistributed; each remains with its provider under that provider's own terms, stated here because they differ.

\textbf{DIAMOR} (ATR, Kyoto) is made available for research purposes only, and its terms require citation of Zanlungo, Ikeda and Kanda (2014) and Gregorj, Y\"ucel, Zanlungo and Kanda (2024); both are cited accordingly. The research-only condition applies to the results reported here and would have to be revisited for any use outside research.

\textbf{J\"ulich} bottleneck and impulse experiments are drawn from the Pedestrian Dynamics Data Archive (Forschungszentrum J\"ulich, doi:10.34735/ped.da), released under CC BY 4.0, which permits reuse with attribution to the source. The ancillary script \texttt{julich\_matched\_state\_robust.py} reproduces the dispersion-controlled and run-level matched-state tests of Section~\ref{sec:julich-matched}. The script \texttt{julich\_operator\_stress\_test.py} reproduces the causal operator/P-ablation, throughput-outcome and direct-operator $\tau$ diagnostics of Section~\ref{sec:operator-stress}. The companion \texttt{julich\_ews\_baselines.py} reproduces the self-only, lag-1 autocorrelation and temporal-variance baseline comparisons reported in the same section. The ancillary \texttt{action\_external\_forcing\_demo.py} reproduces the three-human Physical-AI toy calculation of Section~\ref{sec:operational-object}. The earlier public three-variable matched-state script remains useful as historical provenance; the dispersion-controlled ancillary analysis is the version used for the results reported here.

\textbf{MADRAS/Lyon} trajectories are from the Festival of Lights field dataset of Dufour et al.\ (2025), released under CC BY 4.0 with the data deposited at doi:10.5281/zenodo.13830435.

The Warsaw event recordings (Sections~\ref{sec:probes}, \ref{sec:restricted}) are not redistributed, for three reasons. Participant consent covers research analysis and aggregate reporting, not unrestricted public release of identifiable audiovisual material. The recordings contain identifiable behavior together with contextual and site information, so their release carries privacy and re-identification risk, and de-identification strong enough to remove it would also remove the pose, orientation and interaction geometry the analysis needs. And deployment-specific calibration and instrumentation parameters are part of the proprietary pipeline. Derived non-identifying aggregate outputs and analysis protocols may be shared where participant consent, privacy constraints and proprietary-calibration boundaries permit. The exploratory HRI recording of Section~\ref{sec:rohucad} is likewise not redistributed.

The numerical verification of Theorem~\ref{thm:divergence} in Section~\ref{sec:thm-numerics}, and the computations behind Figures~\ref{fig:theorem-convergence} and~\ref{fig:theorem-geometry}, depend on no restricted data: they are generated entirely from the constructed Jacobian families described there (the frozen family, the pointwise robustness check, the non-normal cross-diffusion family, and the Jordan negative control) and are reproducible from the stated construction with the scripts \texttt{verify\_theorem.py} and \texttt{make\_figures.py}, provided as ancillary files with this preprint.

For the empirical results, we distinguish two cases. Results obtained from exact, seed-fixed reruns, in particular the DIAMOR group-discrimination numbers and the bearing-kernel negative result (Section~\ref{sec:diamor}), are reproducible given access to the source corpus and the corresponding analysis code. Results reported from earlier internal analyses whose exact provenance has not been reconstructed to that standard are labeled \textsc{Candidate} in the ledger of Section~\ref{sec:master-table} for that reason, and are not claimed here as independently reproducible. No result in this paper should be treated as reproducible unless the corresponding analysis code is public.

\section*{Acknowledgments}

The author thanks Prof. Dr. Yulia Sandamirskaya (ZHAW Zurich University of Applied Sciences, Institute of Computational Life Sciences, Neuromorphic Computing) for scientific advice on this work, and Dr. Viktor Malyutin (postdoctoral researcher, Institute of Mathematics, National Academy of Sciences of Belarus) for mathematical discussion relevant to the spectral and stochastic-analysis arguments of Sections~\ref{sec:dynamics}--\ref{sec:theorem}.

% =====================================================================

\end{document}